\documentclass[journal]{IEEEtran}
\usepackage{subfigure}
\usepackage[linesnumbered, ruled, vlined]{algorithm2e}
\usepackage{graphicx}
\usepackage{amsmath} % assumes amsmath package installed
\usepackage{amssymb}  % assumes amsmath package installed
\usepackage{mathrsfs}  
\usepackage{mathtools}
\usepackage{amsthm}
\usepackage{caption}
\usepackage{color}
\usepackage{bm}
\usepackage{booktabs}
\usepackage{multirow}
\usepackage{cite}

\usepackage[implicit=false]{hyperref}

\theoremstyle{plain}
\newtheorem*{problem*}{Problem}

\begin{document}
 
\title{NR-RRT: Neural Risk-Aware Near-Optimal Path Planning in Uncertain Nonconvex Environments}

\author{Fei Meng$\,$\href{https://orcid.org/0000-0001-9225-040X}{\includegraphics[scale=0.08]{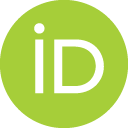}}$\,$,
	Liangliang Chen$\,$\href{https://orcid.org/0000-0002-9594-640X}{\includegraphics[scale=0.08]{figures/ORCIDiD.png}}$\,$,
	Han Ma$\,$\href{https://orcid.org/0000-0003-1960-5432}{\includegraphics[scale=0.08]{figures/ORCIDiD.png}}$\,$,
	Jiankun Wang*$\,$\href{https://orcid.org/0000-0001-9139-0291}{\includegraphics[scale=0.08]{figures/ORCIDiD.png}}$\,$,~\IEEEmembership{Senior Member,~IEEE},\\
	Max Q.-H. Meng*$\,$\href{https://orcid.org/0000-0002-5255-5898}{\includegraphics[scale=0.08]{figures/ORCIDiD.png}}$\,$,~\IEEEmembership{Fellow,~IEEE}
	
	\thanks{This work was supported in part by the Hong Kong RGC TRS grant T42-409/18-R, Hong Kong RGC CRF grant C4063-18G, Hong Kong Health and Medical Research Fund (HMRF) under Grant 06171066,  and Hong Kong RGC GRF grants \#14211420 awarded to Max Q.-H. Meng. \textit{(Corresponding authors: Jiankun Wang and Max Q.-H. Meng.)}}
	
	\thanks{F. Meng and H. Ma are with the Department of Electronic Engineering, The Chinese University of Hong Kong, Hong Kong (e-mail: \{feimeng, hanma\}@link.cuhk.edu.hk).}% <-this % stops a space
	\thanks{L. Chen is with the School of Electrical and Computer Engineering, Georgia Institute of Technology, Atlanta, GA 30332 USA (e-mail: liangliang.chen@gatech.edu).}% <-this % stops a space
	\thanks{J. Wang is with Shenzhen Key Laboratory of Robotics Perception and Intelligence, and the Department of Electronic and Electrical Engineering of the Southern University of Science and Technology, Shenzhen, China (e-mail: wangjk@sustech.edu.cn).} 
	\thanks{Max Q.-H. Meng is with Shenzhen Key Laboratory of Robotics Perception and Intelligence, and the Department of Electronic and Electrical Engineering, Southern University of Science and Technology, Shenzhen 518055, China, on leave from the Department of Electronic Engineering, The Chinese University of Hong Kong, Hong Kong, and also with the Shenzhen Research Institute of The Chinese University of Hong Kong, Shenzhen 518057, China (e-mail: max.meng@ieee.org).} % <-this % stops a space
	
	\thanks{Color versions of one or more of the figures in this article are available online at http://ieeexplore.ieee.org.} 
	
	\thanks{Digital Object Identifier} 
}

\markboth{Journal of \LaTeX\ Class Files,~Vol.~6, No.~1, January~2007}%
{Shell \MakeLowercase{\textit{et al.}}: Bare Demo of IEEEtran.cls for Journals}

\maketitle

\begin{abstract}
Balancing the trade-off between safety and efficiency is of significant importance for path planning under uncertainty.
Many risk-aware path planners have been developed to explicitly limit the probability of collision to an acceptable bound in uncertain environments.
However, convex obstacles or Gaussian uncertainties are usually assumed to make the problem tractable in the existing method.
These assumptions limit the generalization and application of path planners in real-world implementations.
In this article, we propose to apply deep learning methods to the sampling-based planner, developing a novel risk bounded near-optimal path planning algorithm named neural risk-aware RRT (NR-RRT).
Specifically, a deterministic risk contours map is maintained by perceiving the probabilistic nonconvex obstacles, and a neural network sampler is proposed to predict the next most-promising safe state.
Furthermore, the recursive divide-and-conquer planning and bidirectional search strategies are used to accelerate the convergence to a near-optimal solution with guaranteed bounded risk.
Worst-case theoretical guarantees can also be proven owing to a standby safety guaranteed planner utilizing a uniform sampling distribution.
Simulation experiments demonstrate that the proposed algorithm outperforms the state-of-the-art remarkably for finding risk bounded low-cost paths in seen and unseen environments with uncertainty and nonconvex constraints.
\end{abstract}

\def\abstractname{Note to Practitioners}
\begin{abstract}
	This article is motivated by developing an efficient risk-aware path planner that can quickly find risk bounded solutions in uncertain nonconvex environments for practical applications, such as surgical navigation and delivery in crowded squares.
	Sampling-based planning approaches such as rapidly-exploring random tree (RRT) and its variants are popular for their good performance in exploring the state space.
	However, it is quite time-consuming to look for risk bounded paths in uncertain environments, especially under nonconvex and non-Gaussian constraints.
	The initial paths are often of poor quality as well.
	Therefore, we propose the NR-RRT algorithm to rapidly find near-optimal solutions with guaranteed bounded risk.
	It utilizes an informed bidirectional search strategy after having past experiences in those challenging environments.
	NR-RRT can be applied in not only seen but also unseen uncertain scenarios.
	However, the algorithm cannot handle entirely unseen environments that contain new or additional obstacles.
	In future research, we will address the problem of planning under robot model uncertainty.
\end{abstract}


\IEEEpeerreviewmaketitle

\section{Introduction}
%1. PP under uncertainty是什么，目的，期望结果、常用方法
%2. SOTA方法及缺点
%最近神经网络趋势，我们过去的做法
%3. 我们这次的结构做法和改进。take uncertainty into account、unseen、nonconvex去克服缺点
%4. 贡献
%5. 文章结构
\IEEEPARstart{R}{obots} are often required to navigate safely under sensing uncertainties in many practical applications \cite{barbosa2021risk}.
In such circumstances, robots need to plan safe trajectories prior to execution where the probability of collision with uncertain obstacles is limited to a user-specified bound \cite{da2019collision}.
To this end, a desirable risk-aware path planning approach should have the following critical features: 1) completeness and optimality guarantee---a path will be eventually found if one exists, and its cost is the lowest, 2) adaptation to environment constraints---the algorithm can adapt to diverse uncertain environments where the obstacles can be convex or not and have arbitrary probabilistic uncertainties, and effectively generate paths that satisfy any user-preferred risk level, 3) computational efficiency---time and memory consumed to find a solution should be as less as possible.
At present, sampling-based motion planners (SBMPs), such as probabilistic roadmap (PRM) \cite{kavraki1996probabilistic} and rapidly-exploring random tree (RRT) \cite{lavalle1998rapidly}, have become prominent because they can quickly find a solution and guarantee probabilistic completeness \cite{lavalle2006planning}.
\begin{figure}[htbp]
	\centering
	\includegraphics[width=0.48\textwidth]{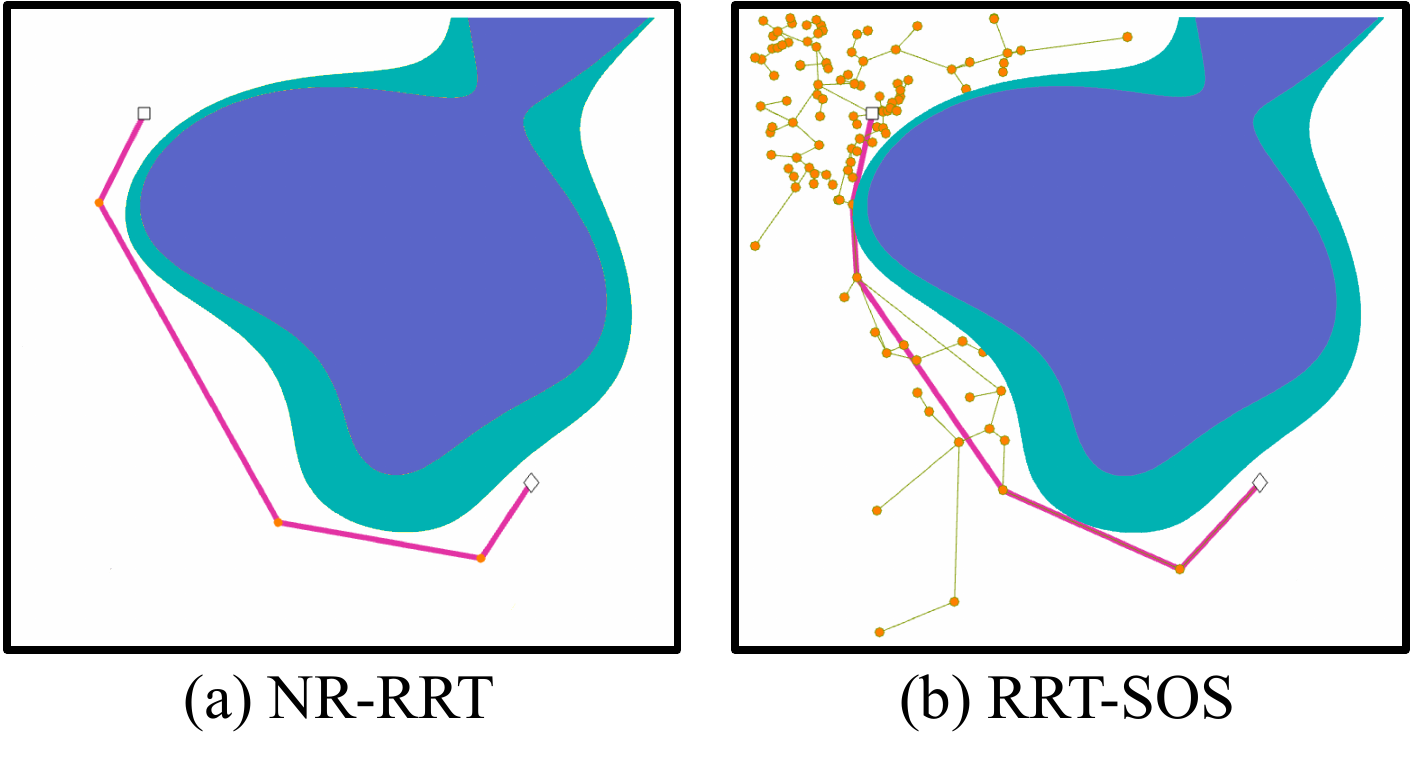}
	\caption{NR-RRT can find comparable risk bounded near-optimal path solutions in uncertain nonconvex environments with much higher computation efficiency after exploiting learned distributions compared to RRT-SOS \cite{jasour2021convex}.}
%		  that has to expand their planning spaces through the exhaustive search before ﬁnding a similarly optimal path.}
	\label{introduction}
\end{figure}
However, the pre-construction of a roadmap makes the PRM-based approaches impractical for online planning.
In contrast, RRT algorithm iteratively builds a rapidly-exploring tree to connect the robot's collision-free states.
The primary solution is obtained by querying the constructed graph after the initial and goal states are contained in the tree.
Despite RRT and its variants guarantee to find a solution, they often fail to find the shortest one.
RRT* \cite{karaman2011sampling}, an optimal variant of RRT, was proposed to additionally keep the cost of the solution decreasing until being the lowest, i.e., it guarantees asymptotic optimality.
Unfortunately, it not only needs to assume perfect measurement information but also takes a large amount of time to find the optimal solution.

Recently, some references have extended the SBMPs to systems with noisy observations to obtain risk bounded trajectories.
For example, Axelrod et al. \cite{axelrod2018provably} modified RRT by verifying safety certificates regarding polytopes with Gaussian distributed faces to generate safe trajectories in uncertain circumstances.
However, most existing sampling-based risk-aware planning methods are limited to either convex obstacles or Gaussian uncertainties \cite{luders2010chance, axelrod2018provably,agha2014firm, ho2022gaussian}.
%To mitigate the issue, Summers et al. \cite{summers2018distributionally} utilize moment-based approaches in motion planning problem that has convex obstacles obeying non-Gaussian probability distributions.
Although the probabilistic location, size, and geometry of uncertain nonconvex obstacles can be explicitly described according to the prior knowledge of environments, dealing with complex and nonconvex constraints is still intractable \cite{jasour2021convex}.
Jasour et al. \cite{jasour2019risk} transformed the uncertain environment into the deterministic information map named risk contours map through moment-based methods, no matter what probabilistic distributions over the uncertain parameters of the obstacles are. 
The map's safe regions also vary with different predefined risk levels.
Then, they proposed RRT-SOS, which constructs an exploring-tree graph in the risk contours map. 
By introducing the sum of squares (SOS) techniques, the algorithm can provide safety guarantees for the edges of a tree without the need for time discretization \cite{jasour2021convex}.
However, this risk-aware path planning algorithm suffers from a heavy computation burden.

The excellent performance achieved by combing traditional SBMPs and machine learning techniques has caught the attention of the motion planning community.
In our previous work \cite{wang2020neural}, an efficient learning-based optimal path planner was developed, the Neural RRT*, which significantly reduces the planning time.
It guides the sampling process of RRT* according to a nonuniform probability distribution that is learned from the expert paths generated by the A* algorithm \cite{hart1968formal}.
In addition, Qureshi et al. \cite{qureshi2020motion} proposed the deep neural network-based bidirectional iterative motion planning method called MPNet to generate collision-free paths in real-time.
By taking advantage of latent space encoded from the workspace, MPNet can explicitly generate the deeply informed samples for the SBMPs in both seen and unseen scenarios of certainty. 
Adding lazy states contraction and bidirectional search strategies additionally contributes to high-performance planning.
However, developing an efficient learning-based risk-aware path planner in uncertain nonconvex environments is still an open problem.

\begin{figure}[htbp]
	\centering
	\includegraphics[width=0.48\textwidth]{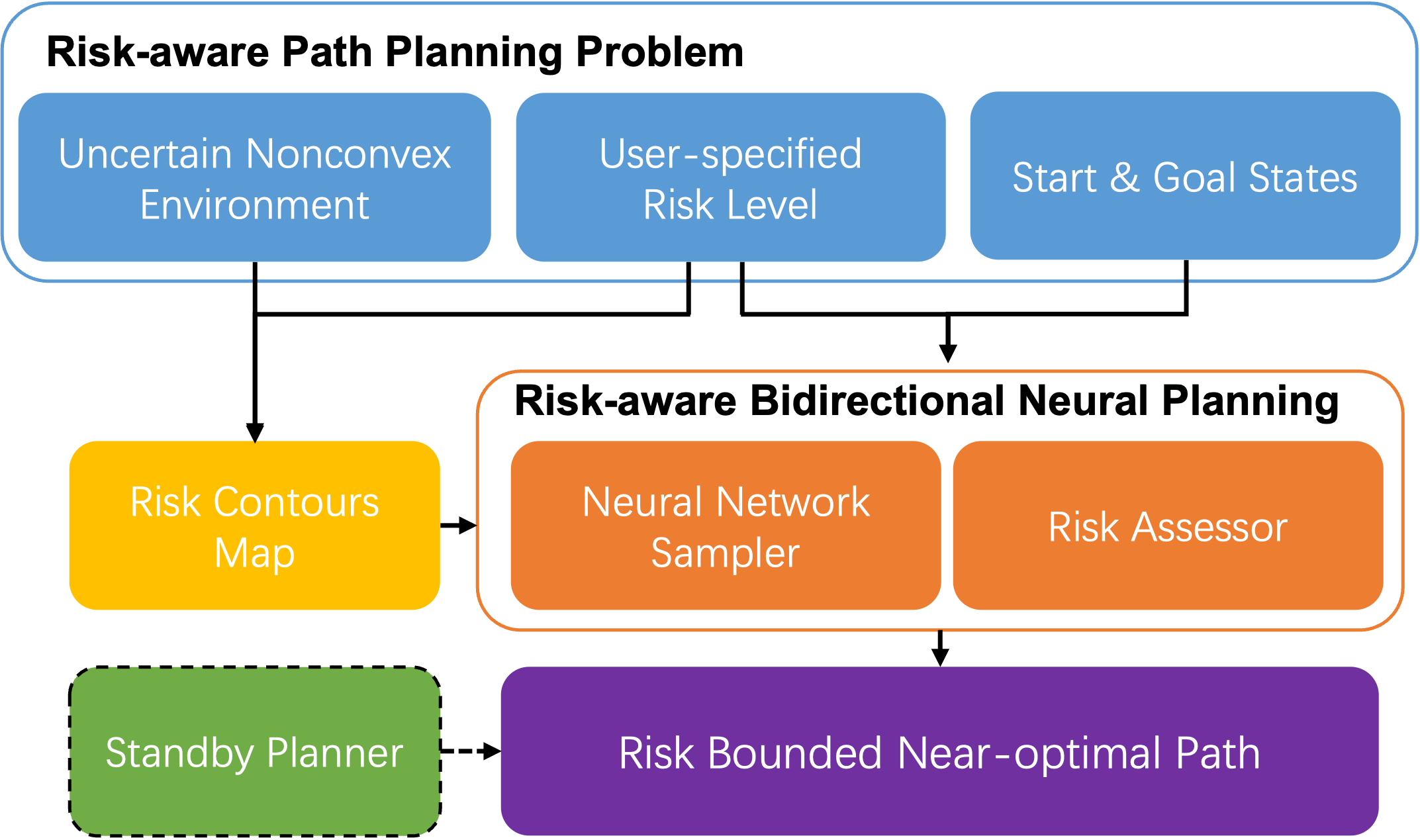}
	\caption{Block diagram illustrating the pipeline of NR-RRT algorithm.}
\label{fig:flowchat}
\end{figure}

To fill this gap, we propose a neural risk-aware path planning method, neural risk-aware RRT (NR-RRT), to find risk bounded near-optimal solutions in uncertain nonconvex environments through learning from RRT-SOS algorithm.
The pipeline of our algorithm is illustrated in Fig. \ref{fig:flowchat}.
A risk-aware path planning problem is formulated given an uncertain nonconvex environment, a user-specified risk level, and the start and goal states.
To capture the information of diverse uncertain nonconvex obstacles, we record their probabilistic locations, sizes, geometries, and the demand of risk level in a risk contours map.
Then, we train a neural network sampler consisting of an encoder and inference network.
The encoder embeds the features of the risk contours map.
At the same time, the inference network learns from abundant expert demonstration paths generated by RRT-SOS to predict the risk bounded node.
The sampler will iteratively output the next informed state that is probably to be contained in the resulting near-optimal solution.
Next, a risk-aware bidirectional neural planning strategy utilizes the sampler to extend two trees incrementally and employs a risk assessor to verify every edge's safety.
This strategy will be recursively called until a fixed number of iterations.
Once the trees are connected, the risk bounded near-optimal path is found.
In case it cannot be found within a limited period, the algorithm calls a standby path planner that guarantees completeness to continue.

The main contribution of this article is threefold.
Firstly, we propose a neural network sampler to iteratively generate the risk bounded and informed states in uncertain nonconvex environments.
Secondly, we propose a computationally efficient risk-aware path planning approach, NR-RRT, which can find the risk-bound near-optimal path solutions rapidly.
Thirdly, our method can generalize to unseen uncertain nonconvex scenarios different from those in the training dataset.
Compared with \cite{qureshi2020motion} and our previous method \cite{wang2020neural}, NR-RRT has improved in three aspects: 1) not only considers risk level, clearance called in \cite{wang2020neural} but also takes explicit probabilistic uncertainties of obstacles into account, 2) can handle path planning problems in both uncertain convex and nonconvex environments, 3) guarantees a bounded probability of collision in such environments.
In addition, NR-RRT can obtain comparable results with significantly reduced computational costs compared to RRT-SOS in \cite{jasour2021convex}.

The remainder of the article is organized as follows. 
We review the main piece of planning methods that are related to this topic in Section \ref{sec:related}.
We formulate the risk-aware path planning problem in uncertain nonconvex scenarios and introduce preliminary in Section \ref{sec:preliminary}, and then present the details of NR-RRT algorithm in Section \ref{sec:nurrt}.
We conduct simulation experiments and analyze the results in Section \ref{sec:experiments}.
Section \ref{sec:discuss} proves worst-cases theoretical guarantees for NR-RRT and discuss its asymptotic optimality.
Finally, we draw conclusions and present consideration of future works in Section \ref{sec:conclu}.

\section{Related Work}
\label{sec:related}
%Many ways have been proposed to overcome the aforementioned limitations of RRT-based algorithms such as bidirectional tree growth \cite{kuffner2000rrt, ma2022bi}, batch sampling \cite{janson2015fast}, and heuristically biasing RRT generation \cite{gammell2015batch,gammell2018informed}, etc.
A risk-aware path planning algorithm aims to find a feasible path connecting given start and goal states, if one exists, in an uncertain environment; meanwhile, the probability of collision with uncertain obstacles for the solution is no larger than a user-specified bound \cite{da2019collision}.
%Many motion planning approaches are designed to look for such solutions for robots under uncertainty \cite{quintero2021robust, luders2013robust,suresh2021planning,agha2014firm,lenz2015heuristic}.
To this end, SBMPs have been extended to look for such solutions under measurement uncertainty \cite{janson2018monte,cannon2017chance,ho2022gaussian}. 
For instance, Janson et al. \cite{janson2018monte} proposed a Monte Carlo-based motion planner to find safe trajectories in which the probability of collision is estimated by sampling. 
However, there are no analytical bounds of the collision risk of the trajectory. 
An alternative for planning under uncertainty is the chance-constraint strategy \cite{blackmore2011chance,johnson2021chance,sun2016stochastic}. 
Instead of maximizing the probability of success, it aims to compute a feasible solution in which the waypoints or segmented trajectories satisfy a minimum probability of collision constraint. 
Johnson et al. \cite{johnson2021chance} developed a motion planner that was capable of reducing the computation complexity in high-dimensional and obstacle-ﬁlled spaces.
Furthermore, performance improvements can be gained by combining stochastic optimal control with traditional SBMPs \cite{luders2010chance,sun2015high,van2011lqg,jaillet2011eg}. 
The linear-quadratic Gaussian motion planning algorithm (LQG-MP) \cite{van2011lqg} incorporated a local controller into the extensions procedure of RRT to deal with noisy sensing. 
Its extended version environment-guided RRT \cite{jaillet2011eg} estimated path quality and guided sampling toward safer parts of the state-space. 
Planning under uncertainty can also be modeled as a Markov decision process (MDP) \cite{alterovitz2007stochastic} if the fully observable system states exist, otherwise as a partially observable Markov decision process (POMDP) \cite{van2012motion,cai2021hyp,lee2020magic}.
For example, Van et al. \cite{van2012motion} proposed the belief iterative LQG to obtain a time-varying affine feedback controller by approximating the quadratic value function over a belief space.
It has all feasible probability distributions over the robot state space. 
However, the methods above were built upon strong assumptions, e.g., convex obstacles and sensor noise sampled from a Gaussian distribution, which are limited in more general cases such as nonconvex obstacles and non-Gaussian measurement noises.
Note that complex and nonconvex constraints are often imposed in planning problems due to the safety reason \cite{richards2002aircraft}.
%and other works \cite{pairet2021online,quintero2021robust,christensen2021closed} 
A new arisen branch of the planning method can address non-Gaussian uncertainties \cite{summers2018distributionally}, nonconvex planning problem \cite{da2019collision}, or combination of them \cite{jasour2021convex} when sensing uncertainty dominates.
However, the computational costs of these methods are still prohibitive because they need plenty of samples or multiple iterations to find a solution.

In recent years, reinforcement learning (RL) and deep learning (DL) have emerged as promising motion planning tools \cite{https://doi.org/10.1049/csy2.12020}.
The advanced version of RL, called deep RL (DRL), is shown capable of effectively planning under uncertainty through incorporating traditional RL with deep neural networks \cite{srinivas2018universal,tamar2016value}.
Faust et al. \cite{faust2018prm} trained multiple RL agents on the smallest map in simulation with sensing noise to learn robust near point-to-point navigation policies.
Then, the PRM-based global planner uses the best policy to construct roadmaps in the robot's C-space where all configurations along the path are safe, realizing a long-range motion planner.
To learn resilient actions for navigation in unseen uncertain environments, Fan et al. \cite{fan2020learning} presented an uncertainty-aware predictor and policy network to gain the uncertain information of circumstances and learn the desirable behaviors, respectively.
Although these methods can generalize to new environments or tasks, they may suffer from weak rewards or require discretization of the state space.

The function approximation ability of deep neural networks can be employed to speed up the convergence of SBMPs to the optimal solution \cite{ichter2018learning,ichter2019robot}.
One main category is to learn bias sampling heuristic to guide the sampling process, namely, to learn the promising probability distribution of the optimal solution \cite{wang2020neural,ma2022enhanced}, while another is to use neural networks to embed a planner \cite{qureshi2020motion,ichter2019robot}.
For example, Zhang et al. \cite{zhang2021generative} proposed a generative adversarial network model to compute the most promising area on a map such that the sampling process can be guided.
%A sampling distribution was learned implicitly through learning a rejection sampling policy \cite{zhang2018learning}. 
%It is then used to judge whether the uniform samples are eligible while creating the SMP graph.
A method that constructs the SBMP graph and conducts the collision-checking procedure both in learned latent spaces was proposed in \cite{ichter2019robot}.
Although these neural planners have achieved remarkable performance in deterministic convex environments, realizing rapid path planning in uncertain nonconvex scenarios remains challenging.

\section{Preliminary}
\label{sec:preliminary}
\subsection{Problem Definition}
\label{sec:problem define}
Let $\mathcal{X}\in\mathbb{R}^{n_x}$ be an uncertain environment containing safe space $\mathcal{X}_\text{safe}\subset \mathcal{X}$ and static uncertain obstacles $\mathcal{X}_{{\rm{obs}}_i}(\omega_i)\subset \mathcal{X},\, i=1,\dots,n_{o}$, where $\omega_i\in \mathbb{R}^{n_\omega}$ represent probabilistic parameters with given probability distributions. 
All obstacles can be convex or nonconvex, and each obstacle may have an uncertain size, location, or geometry. 
We represent them in terms of polynomials in $\bm{x}\in\mathcal{X}$ as below:
\begin{equation}
	\mathcal{X}_{{\rm{obs}}_i}(\omega_i)=\left\{\bm{x}\in\mathcal{X}:\mathcal{P}_i(\bm{x},\omega_i)\geq0\right\},\,i=1,\dots,n_{o}, \label{eq:u_obs}
\end{equation}
where $\mathcal{P}_i:\mathbb{R}^{n_x}\rightarrow\mathbb{R}$ denotes the known polynomial uncertain obstacles.

Take an ellipse-shaped uncertain convex obstacle as an example,  and a nonconvex obstacle is similar. The polynomial of it is $\mathcal{P}:\{(x,\,y):\,\omega^2-x^2/2-y^2\geq0,\,\omega\sim\mathcal{N}(0,1)\}$. Its major and minor axes are subjected to a Gaussian distribution of random variable $\omega$. Note that the type and dimension of the probability distribution should depend on the realistic situation.

Given a safe initial state $\bm{x}_{\rm{init}}$, goal state $\bm{x}_{\rm{goal}}$, and an uncertain environment, we intend to find the shortest and risk bounded path $\bm{\pi}=\left[\bm{x}_{\rm{init}},\dots,\bm{x}_{\rm{goal}}\right]\,:\,[0,T]\rightarrow\mathcal{X}_\text{safe}$ such that  $\bm{\pi}(0)=\bm{x}_{\rm{init}}$ and $\bm{\pi}(T)\in\mathcal{X}_{\rm{target}}(\bm{x}_{\rm{goal}})$, where $\mathcal{X}_{\rm{target}}(\bm{x}_{\rm{goal}})=\{\bm{x}\in\mathcal{X}_{\text{safe}}\,|\,\|\bm{x}-\bm{x}_{\rm{goal}}\|<r\}$ is the target region commonly used in practice. Meanwhile, the probability of collision with uncertain obstacles is no greater than $\Delta$, where $\Delta\in[0,1]$ is a predefined constant representing the acceptable risk level.

Further, the risk-aware optimal path planning is defined mathematically as follows:
\begin{align}
	&\min \limits_{\bm{\pi}:[0,T]\rightarrow\mathcal{X}_\text{safe}} \int_{0}^{T} \left\|\dot{\bm{x}}(t)\right\|_2^2dt \label{eq:cost}\\  
	&\text{s.t.}\, \bm{x}(0) = \bm{x}_{\rm{init}},\,\bm{x}(T) \in \mathcal{X}_{\rm{target}}, \label{eq:initgoal}\\
	&\text{Prob}(\bm{x}(t)\in\mathcal{X}_{{\rm{obs}}_i}(\omega_i))\leq\Delta,\, \forall t \in [0, T]\,|_{i=1}^{n_{o}}\,,	\label{eq:UPC}
\end{align}
where \eqref{eq:cost} denotes the cost function regarding the length of trajectory $\bm{\pi}$ measured by $\mathscr{L}_2$ norm, and
\eqref{eq:UPC} are the collision risk constraints in terms of probability for the trajectory $\bm{\pi}$. 

\subsection{Risk Contours Map}
We introduce a cutting-edge mapping method, called risk contours map \cite{jasour2019risk}, to describe the probabilistic information of uncertain obstacles.
This map depicts the relatively safe region in the uncertain environment where the probability of collision with the uncertain obstacles is no greater than a given risk level $\Delta$. 
Specifically, considering one uncertain obstacle $\mathcal{X}_{{\rm{obs}}}(\omega)$ in \eqref{eq:u_obs}, we define the $\Delta$-risk contour $\mathcal{C}_r^{\Delta}$ as the following set of states:
\begin{equation}
	\mathcal{C}_r^{\Delta}\coloneqq\left\{\bm{x}\in\mathcal{X}:\text{Prob}\left(\bm{x}\in\mathcal{X}_{{\rm{obs}}}(\omega)\leq\Delta\right)\right\}.
	\label{eq:contour}
\end{equation}
%Note that the contour is constructed with respect to only one uncertain obstacle, and \eqref{eq:contour} gives the feasible set of optimization under \eqref{eq:UPC} by assigning which one obstacle to be considered.

Then, to transform the original probabilistic optimization problem \eqref{eq:cost} into a solvable one, a deterministic constraint in terms of the set of safe states is approximated as follows \cite{jasour2021convex}:
\begin{equation}
	\widehat{\mathcal{C}}_r^{\Delta}\!=\!\left\{\!\bm{x}\!\in\!\mathcal{X}\!\!:\!\!\!\,\frac{\mathbb{E}[\mathcal{P}^2(\bm{x}, \omega)]\!\!-\!\! \mathbb{E}[\mathcal{P}(\bm{x},\omega)]^2}{\mathbb{E}[\mathcal{P}^2(\bm{x}, \omega)]}\!\leq\!\Delta,
	\mathbb{E}[\mathcal{P}(\bm{x},\omega)]\!\leq\!0\!
	\right\}
	\label{eq:hat contour}
\end{equation}
where $\mathcal{P}(\bm{x}, \omega)$ is the known polynomial of the uncertain obstacle $\mathcal{X}_{{\rm{obs}}}(\omega)$.
$\mathbb{E}[\mathcal{P}(\bm{x},\omega)]$ and $\mathbb{E}[\mathcal{P}^2(\bm{x}, \omega)]$ are polynomials in terms of states $\bm{x}$ and the moments of probabilistic parameter $\omega$. In other words, the coefficients of $\bm{x}$ in them are computed by using the moment of $\omega$ and the coefficients of $\mathcal{P}(\bm{x}, \omega)$.

Note that the set $\widehat{\mathcal{C}}_r^{\Delta}$ is a rational polynomial-based inner approximation of the original risk contour $\mathcal{C}_r^{\Delta}$. 
Thus, we have a deterministic constraint in terms of the set of risk bounded states $\bm{x}\in\widehat{\mathcal{C}}_r^{\Delta},\,\forall t\in[0,T]$.
The resulting path $\bm{\pi}$ comprising of these states is guaranteed to have a no greater than $\Delta$ chance of collisions. 
For more details, the readers are referred to \cite{jasour2021convex}. 
Finally, we have a deterministic polynomial optimization problem with risk contours as follows:
\begin{align}
	&\min \limits_{\bm{\pi}:[0,T]\rightarrow\mathcal{X}_\text{safe}} \int_{0}^{T} \left\|\dot{\bm{x}}(t)\right\|_2^2dt \label{eq:deter_cost}\\  
	&\text{s.t.}\, \bm{x}(0) = \bm{x}_{\rm{init}},\,\bm{x}(T) \in \mathcal{X}_{\rm{target}}, \label{eq:deter_initgoal}\\
	&\bm{x}(t)\in\widehat{\mathcal{C}}_{r_i}^{\Delta},\,\forall t \in [0, T]\,|_{i=1}^{n_{o}}.	\label{eq:deter_UPC}
\end{align}
%It can be used to solve a trajectory optimization problem under probabilistic constraints online as it is an analytical method. 
\subsection{RRT-SOS}
RRT algorithm is known for rapid space exploration and can always find a solution, if one exists, as the number of samples keeps increasing.
It first performs sampling from the state space according to some sampling distribution, given a start state as the root vertex of a searching tree.
%After measuring the distances between the sampled node $x_\text{sample}$ and near node(s) in the tree, 
Then, it selects the nearest vertex $x_{\rm{nearest}}$ and tries to connect it through an extension function.
If the connection is collision-free after performing collision detection, $x_{\rm{new}}$ adjusted from $x_\text{sample}$ will be a new vertex and added to the vertex set of the searching tree.
The pair $(x_{\rm{nearest}},\,x_{\rm{new}})$ will also be included in the edge set.
Repeating the procedures above until a state in the target region $\mathcal{X}_{\rm{target}}$  is contained in the tree, and then the algorithm returns the solution.
However, the collision detection needs to be modified since we have probabilistic polynomial information $\mathcal{P}_i(\bm{x},\omega_i) $ rather than certain polynomial information $\mathcal{P}_i(\bm{x})$ of the obstacles in uncertain environments.
We need to make sure the probability of collision for the connected path is bounded by $\Delta$ instead of zero.
In other words, the paths should be risk bounded rather than purely collision-free due to environmental uncertainty.

For this purpose, Jasour et al. \cite{jasour2021convex} proposed RRT-SOS algorithm, which uses the safety sum of squares (SOS) condition as follows to check the collision risk of the trajectory between $x_{\rm{nearest}}$ and $x_{\rm{new}}$ in an uncertain environment.
$\mathcal{S}=\{\bm{x}:g_k(x)\geq0,\,k=1,\dots,l\}$ is defined as the feasible point set of \eqref{eq:deter_cost}, where $g_k(\cdot)\geq0,\,k=1,\dots,l$ are polynomial constraints of all the inner approximations of risk contours $\widehat{\mathcal{C}}_{r_i}^{\Delta},\,i=1,\dots,n_{o}$. The trajectory $\bm{x}(t)$ between two vertices satisfies the constraint \eqref{eq:deter_UPC} over the time interval $[t_1,\,t_2]$, iff polynomials $g_k(\bm{x}(t)),\,k=1,\dots,l$ has the SOS form below:
\begin{equation}
	g_k(\bm{x}(t))=\pi_{0k}(t)+\pi_{1k}(t)(t-t_1)+\pi_{2k}(t)(t_2-t)|_{k=1}^{l},
	\label{eq:sos condition}
\end{equation}
where $\pi_{0k}(t),\,\pi_{1k}(t)$, and $\pi_{2k}(t)$ are SOS polynomials with suitable degrees.

The authors in \cite{jasour2021convex} also initialized a straight line between the start and target vertices to adaptively sample in its neighborhood during the expansion, which relieved the computation burden to some extent.
After finding a primary path that connects the start and goal states, a PRM graph whose nodes and edges satisfy the SOS condition in \eqref{eq:sos condition} is constructed for the solution.
Then, the Dijkstra algorithm, a shortest path-finding method, is performed to reduce the path length of the feasible solution by querying the generated graph.

\section{NR-RRT: A neural risk-aware path planner}
\label{sec:nurrt}
In this section, we first describe how to process the uncertain environment to facilitate the training of our neural networks in Section \ref{sec:dataset}.
The architecture and other details of the proposed neural network sampler are introduced in Sections \ref{sec:Model}. 
NR-RRT algorithm is presented in Section \ref{sec:nrrrt}.

\subsection{Pre-processing}
\label{sec:dataset}
2-D images are commonly used to represent the 2-D space.
An image is usually converted into an occupancy grid map in which zero/one indicates free/occupied space in definite environments. 
However, it is insufficient to present the space information of uncertain obstacles using this kind of map because we cannot be sure whether some grids are occupied or not.
To solve this problem, we construct a risk contours map that replaces the probabilistic grids with deterministic ones in the image.
Specifically, by computing the inner approximation of static risk contour in \eqref{eq:hat contour}, we obtain continuous risk levels in $[0,1]$ for the coordinates in the risk contours map instead of $\{0,1\}$ used for grid-based maps.

\begin{figure}[htbp]
	\centering
	\includegraphics[width=0.48\textwidth]{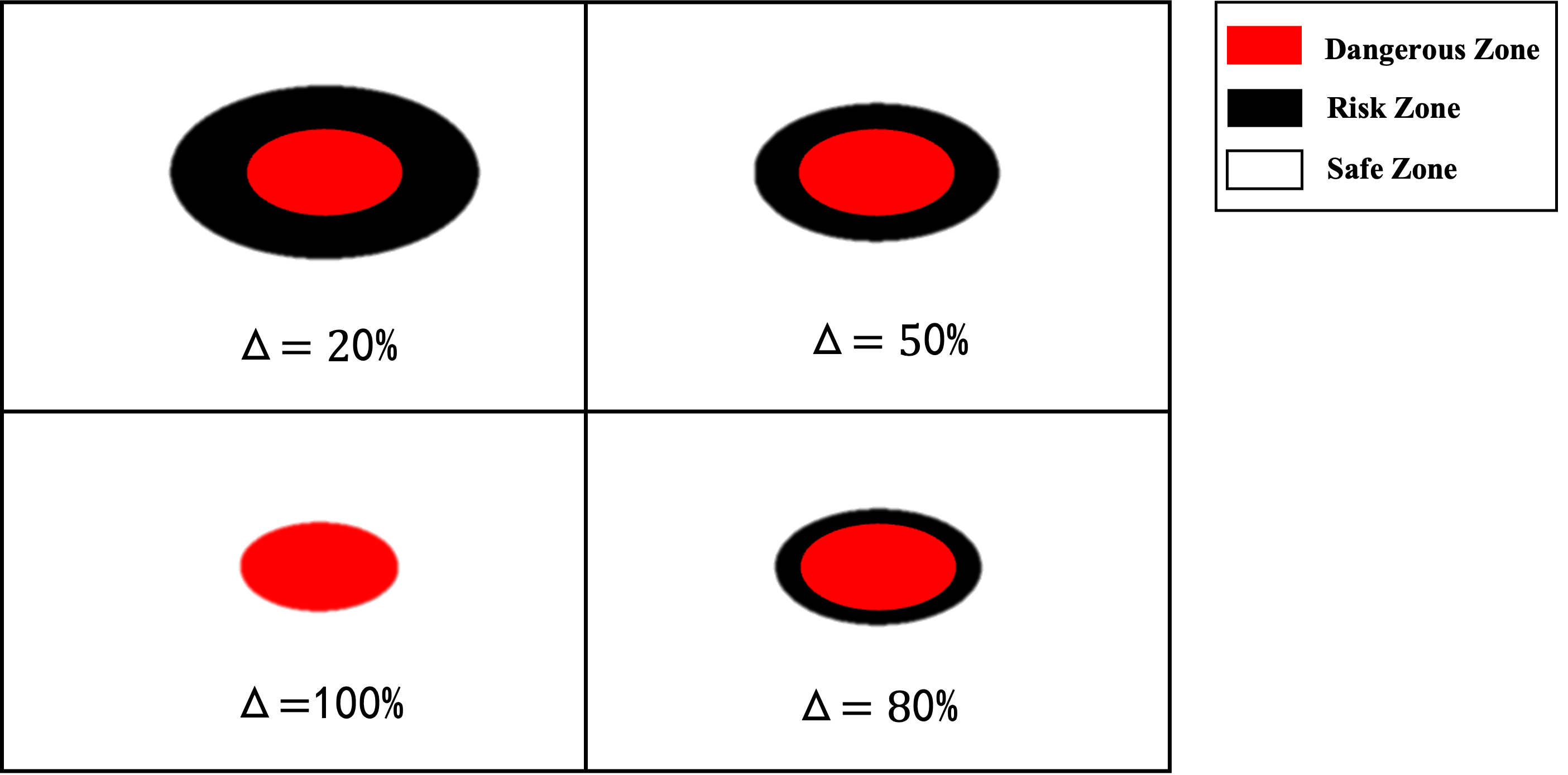}
	\caption{The risk contours maps with different $\Delta$ for an ellipse-shaped uncertain obstacle. $\Delta$ is a predefined acceptable probability of collision. The dangerous zone (red) is fixed. In contrast, the risk zone (black) shrinks as the risk level $\Delta$ increases, as shown in the subfigures clockwise from the left-top, indicating that the farther away from the dangerous zone, the smaller probability of collision.}
	\label{fig:contour}
\end{figure}

We take the 2-D ellipse-shaped uncertain obstacle in Section \ref{sec:problem define} as an example again. 
Given the polynomial $\mathcal{P}(\bm{x},\,\omega)$ and the distribution of uncertain parameter $\omega\sim\mathcal{N}(0,1)$, we can calculate the first and second moments as 
\begin{equation}
	\begin{split}
		&\mathbb{E}[\mathcal{P}(\bm{x},\omega)]=\mathbb{E}[\omega^2-0.5x^2-y^2]\\
		&=-0.5x^2-y^2+\mathbb{E}[\omega^2]=-0.5x^2-y^2+1,\\
		&\mathbb{E}[\mathcal{P}^2(\bm{x},\omega)]=\mathbb{E}[(\omega^2-0.5x^2-y^2)^2]\\
		&=0.25x^4+y^4+x^2y^2-\mathbb{E}[\omega^2](x^2+2y^2)+\mathbb{E}[\omega^4]\\
		&=0.25x^4+y^4+x^2y^2-x^2-2y^2+3.
		\label{eq:firstsecondmoment}
	\end{split}
\end{equation}
We substitute the moments \eqref{eq:firstsecondmoment} into the deterministic constraint $\widehat{\mathcal{C}}_r^{\Delta}$ in \eqref{eq:hat contour}, obtaining two inequalities without the uncertain parameter but symbolic variables, $x$ and $y$.
By substituting the coordinates $(x,y)$ in the environment into the inequalities, we can split the risk contours map into three parts: safe zone, dangerous zone, and risk zone, as shown in Fig. \ref{fig:contour}.
More precisely, the safe zone (white) means the inequalities in \eqref{eq:hat contour} hold for all coordinates in it, while the dangerous zone (red) and risk zone (black) indicate $\mathbb{E}[\mathcal{P}(\bm{x},\omega)]\leq0$ and $\frac{\mathbb{E}[\mathcal{P}^2(\bm{x}, \omega)]- \mathbb{E}[\mathcal{P}(\bm{x},\omega)]^2}{\mathbb{E}[\mathcal{P}^2(\bm{x}, \omega)]}\!\leq\!\Delta$ do not hold, respectively.
In other words, only states in the safe zone have a no greater than $\Delta$ chance of collisions.

Furthermore, the image has a size of $W\times H\times C$, corresponding to width, height, and the number of channels, respectively, and every pixel in it has a value from 0 to 255 in each channel. 
As a result, pixels in the image-based risk contours map have three kinds of values corresponding to three colors of the zones, i.e., the values of pixels are $(0,0,0)$ in the risk zone (black), $(255,255,255)$ in the safe zone (white), and $(255,0,0)$ in the dangerous zone (red).
Note that we do not require discretization of the 2-D environment, and the positions of pixels in the image are not equal to the coordinate values in the risk contours map, which means our method will not suffer from the exponentially increased computation cost brought by the increased map size.
Therefore, we record the continuous probabilistic observations of uncertain obstacles in image $I$.
We also normalize the path data generated by RRT-SOS with respect to the boundaries of the risk contours map to facilitate the training.

\subsection{Neural Network Sampler Architecture}
\label{sec:Model}
\begin{figure*}[htbp]
	\centering
	\includegraphics[width=1\textwidth]{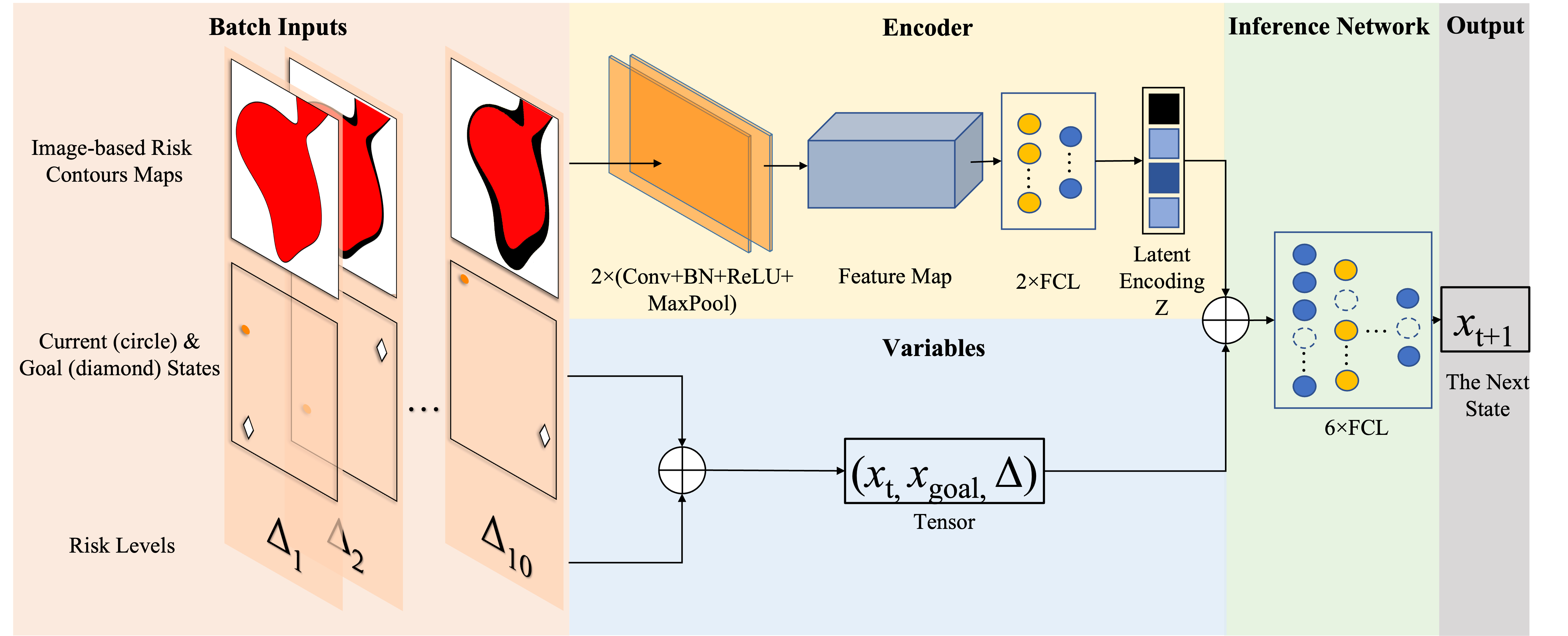}
	\caption{The overall structure of the neural network sampler for informed sampling.}
	\label{fig:model}
\end{figure*}

%Even though RRT-like algorithms are famous for rapid exploring the state space using a sampling manner, it is still potential to reduce the time cost of planning through a sampling distribution learned from expert data.
We design an end-to-end neural network sampler (NNS) consisting of an observation encoder network $\mathcal{E}$ with weights $\theta_e$ and an inference network $\mathcal{I}$ with weights $\theta_i$, which is inspired by MPNet \cite{qureshi2020motion}. 
The architecture of our proposed neural networks is shown in Fig. \ref{fig:model}. 

Firstly, the encoder $\mathcal{E}$ embeds the observation of obstacles in images $I$ into latent features $Z$,
\begin{equation}
	\mathcal{E}(I;\theta_e)\rightarrow Z.
\end{equation}
The fully connected networks treat pixels far apart and those close together in the same way.
It ignores the local features and is not suitable for image data processing.
By contrast, the convolutional neural networks (CNNs) have impressive features such as local connectivity, shared weights, and pooling, leading to their better generalization with lower memory and fewer free parameters \cite{girshick2015fast}.
We employ a CNN as our encoder to automatically extract latent features $Z$ from images $I$.
It has two 2D convolutional layers with the kernel size $3\times3$.
Each is connected with batch normalization (BN), an activation function, the rectified linear unit (ReLU), and a downsampling layer, max pooling, with filters of size $2\times 2$.
The input and output channels of the ﬁrst and second layers are (3, 32) and (32, 128), respectively.
Every convolutional layer takes one image-based map and outputs a low-level feature map.
The ReLU allows the CNN to be trained faster, and the max-pooling layer makes it more robust to variations in the shapes of uncertain obstacles.
The image is then abstracted to the latent features $Z$ with a size of 64 through two fully connected layers (FCLs).

After that, the latent-space embedding $Z$ concatenated with the current state $\bm{x}_t\in\mathcal{X}_{\text{safe}}$, goal state $\bm{x}_{{\rm{goal}}}\in\mathcal{X}_{\text{safe}}$, and risk level $\Delta$ are fed into the inference network $\mathcal{I}$, and it outputs the next state $\hat{\bm{x}}_{t+1}.$ towards the goal state $\bm{x}_{{\rm{goal}}}$,
\begin{equation}
	\mathcal{I}(Z,\bm{x}_t,\bm{x}_{\rm{goal}},\Delta;\theta_i)\rightarrow \hat{\bm{x}}_{t+1}.
\end{equation}
The inference network $\mathcal{I}$ comprises six FCLs, and each of them has Dropouts \cite{srivastava2014dropout} to let the network $\mathcal{I}$ play a stochastic behavior. 
This way increases the probability of exploring the whole state space rather than being trapped in several informed samples, which is better to maintain the completeness guarantee.

We use a mean-squared-error (MSE) loss between the predicted state $\hat{\bm{x}}_{t+1}$ and the label next state $\bm{x}_{t+1}$ for both the encoder $\mathcal{E}$ and inference network $\mathcal{I}$ during training, i.e.,
\begin{equation}
	l(\theta_e,\theta_i)=\frac{1}{N_a}\sum_{p}^{\hat{N}}\sum_{q=0}^{L_p-1}\left\|\hat{\bm{x}}_{p,q+1}-\bm{x}_{p,q+1}\right\|^2,
\end{equation}
where $N_a$ is a positive averaging constant, $\hat{N}$ is the total number of training paths in an environment, and $L_p$ is the length of $p$-th path, respectively.

Therefore, the NNS is recursively called to make the newly generated state $\hat{\bm{x}}_{t+1}$ become the robot's current state $\bm{x}_t$ in the next time step, leading the searching tree to expand incrementally.

\subsection{Online Path Planning}
\label{sec:nrrrt}
This section introduces the key components of NR-RRT, a bidirectional neural risk-aware path planning algorithm. 
The procedures of NR-RRT are summarized in Algorithm \ref{al:nrrrt}.

Now, the neural network sampler in Section \ref{sec:Model} can prompt the tree to expand efficiently. 
The algorithm still needs a \textit{Risk Assessor (RA)} to verify that the risk of collision is bounded during the expansion.
\textit{RA} is used to verify a vertex or edge whether lies in the safe zone, and it returns True if so or False if not.
Given a vertex, we substitute its states and predefined risk level $\Delta$ into the inner approximation of risk contour $\widehat{\mathcal{C}}_r^{\Delta}$. 
As we did to decide which area is the safe zone, the vertex is risk bounded if the inequalities in \eqref{eq:hat contour} hold.
On the other hand, an edge between two vertices is a straight line segment and can be rewritten as a linear polynomial trajectory such as $\{(x,y),x=a_1t+b_1,y=a_2t+b_2\}$, given the two endpoint states.
The sum of squares (SOS) form as \eqref{eq:sos condition} will be verified for the edge using the Spotless package \cite{tobenkin2013spotless}.
As a result, the collision probability along this edge is smaller than or equal to $\Delta$ if it satisfies the SOS condition.
Moreover, we use a hybrid replanning strategy \cite{qureshi2020motion} to speed up the algorithm further and guarantee its completeness.
In principle, given the original planning problem, NR-RRT performs bidirectional neural planning at first (described later) and outputs a coarse global path.
Then, if any edge between two contiguous vertices lies in the risk or dangerous zone, we call \textit{Replan} to replan this edge.
This function takes the non-connectable nodes as a start and goal pair and utilizes the bidirectional neural planning again to recursively generate a local path for them.
If there is no solution available yet, in case some problems are hard, we employ the standby planner that exhibits probability completeness to find it.
Since the primary solution is usually not optimal, we implement the shortcutting process, \textit{Lazy States Contraction (LSC)} \cite{hauser2010fast}, to eliminate the useless vertices in it.
With \textit{Replan} and \textit{LSC}, the algorithm  keeps finding possible critical connectable vertices and removing the unconnectable nodes.
%\begin{figure}[htbp]
%	\centering
%	\includegraphics[width=0.2\textwidth]{./figures/contour.png}
%	\caption{diveconqure}
%	\label{diveconqure}
%\end{figure}

As shown in Algorithm \ref{al:nrrrt}, NR-RRT takes the start and goal states, $\bm{x}_{{\rm{init}}}$ and $\bm{x}_{{\rm{goal}}}$, uncertain obstacles $\mathcal{X}_{{\rm{obs}}_i}$, and user-specified risk level $\Delta$ as inputs.
The encoder $\mathcal{E}$ encodes the image-based risk contours map $I$ into a latent feature map $Z$ (Line 1).
We initialize two trees, i.e., $\mathcal{T}_f=(V_f,E_f)$ to grow forwards from start to goal and $\mathcal{T}_b=(V_b,E_b)$ to grow backwards from goal to start, respectively, and an empty path solution $\bm{\pi}$ (Line 2).
Each tree comprises a vertex set $V\subset\mathcal{X}_{\rm{safe}}$ and an edge set $E\subseteq V\times V$.
We expand trees in an alternating manner and repeat this manner for a fixed number of iterations $N$.
We take the expansion step for forward path $\bm{\pi}_f$ (Lines 5-14) as an example here.
Our method uses \textit{NNS} to generate a sample $\bm{x}_{\rm{sample}}$, given the latent encoding $Z$, and the top nodes, $\bm{x}_f^{end}$ and $\bm{x}_b^{end}$.
Then, $\bm{x}_{\rm{sample}}$ is verified through \textit{RA}, and if the collision risk is bounded, it will be adjusted to $\bm{x}_{\rm{new}}$ and added to the vertex set $V_f$.
It and its nearest vertex in the tree $\bm{x}_{\rm{nearest}}$ are inputted into \textit{Steer} (Line 9).
In \textit{Steer}, we connect two vertices with a straight line and perform \textit{RA} to assess the collision risk for their edge. 
If risk bounded, it is added to the edge set $E_f$.
After each expansion step, the algorithm tries to connect both trees by calling \textit{Steer} again (Line 11).
Once the connection is successful, we obtain a coarse tree $\mathcal{T}$ containing the vertices from $\mathcal{T}_f$ and $\mathcal{T}_b$, and have a primary path solution $\bm{\pi}=\{\bm{x}_{{\rm{init}}},\dots,\bm{x}_{{\rm{goal}}}\}$ by querying the tree $\mathcal{T}$.
If the connection fails, we swap $\mathcal{T}_f$ with $\mathcal{T}_b$ and keep expanding (Line 15).
If a solution $\bm{\pi}$ is found, we reﬁne it with \textit{LCS}.
The \textit{LCS} function will return critical vertices regarding the optimal path.
\textit{RA} is then utilized to check the collision risk of every edge in the global path.
Suppose there is a local path crossing either the dangerous or risk zone.
In that case, we input the failed global path into \textit{Replan}, followed by \textit{LSC}, to generate a near-optimal solution until the number of iterations reaches the threshold $N_j$ (Lines 20-24).
In case the strategy using bidirectional neural planning fails after $N_j$ attempts, the algorithm employs the standby planner, RRT-SOS, which samples from a uniform sampler.
By adjusting the non-connectable nodes as new start and goal states, RRT-SOS attempts to find a path solution while maintaining the theoretical probabilistic completeness (Lines 30-32). 
Eventually, the resulting path is optimized through \textit{LSC} and the collision risk of it is verified through \textit{RA} again (Lines 35-36).

\begin{algorithm}[t]
	\caption{NR-RRT ($\bm{x}_{\rm{init}}$, $\bm{x}_{\rm{goal}}$, $\mathcal{X}_{\rm{obs}_i}$, $\Delta$)}
	\label{al:nrrrt}
%	\KwIn{$\bm{x}_{\rm{init}} \in \mathcal{X}_{\rm{safe}}$, $\bm{x}_{\rm{goal}} \subset \mathcal{X}_{\rm{safe}}$, $\mathcal{X}_{\rm{obs}}$, $\Delta$} 
%	\KwOut{$\bm{\pi}$} 
	
	 $\mathcal{E}(I(\mathcal{X}_{\rm{obs}_i}))\rightarrow Z$\;
	 $\mathcal{T}_f=(V_f,E_f)$, $\mathcal{T}_b=(V_b,E_b)$, $\bm{\pi}\gets \varnothing$\;
	 $V_f\gets \{\bm{x}_{\rm{init}}\}$, $V_b\gets \{\bm{x}_{\rm{goal}}\}$\;
	\For{$i=0,\dots, N$}
	 {$\bm{x}_{\rm{sample}} \gets NNS \left(Z, V_f^{end}, V_b^{end}\right)$\;
	\While{$RA(\bm{x}_{\rm{sample}},\Delta)=False$}
	{$\bm{x}_{\rm{sample}} \gets NNS \left(Z, V_f^{end}, V_b^{end}\right)$\;}
	{$V_f\gets V_f\cup\bm{x}_{\rm{new}}$\;}
	
	\If {$\textit{Steer}(\bm{x}_{\rm{nearest}}, \bm{x}_{\rm{new}})$}
	{$E_f\gets E_f\cup(\bm{x}_{\rm{nearest}}, \bm{x}_{\rm{new}})$\;}
	
	\If{$\textit{Steer}(V_f^{end}, V_b^{end})$}{
	{$\mathcal{T}\gets \rm{concatenate}(\mathcal{T}_f, \mathcal{T}_b)$}\;
	$\bm{\pi}=\{\bm{x}_{\rm{init}},\dots,\bm{x}_{\rm{goal}}\}$\;
	$Break$\;}
	
	SWAP($\mathcal{T}_f, \mathcal{T}_b$)\;
	}
	$\bm{\pi}\gets LSC (\bm{\pi})$\;
	\If{$RA(\bm{\pi},\Delta)$}{
	\textbf{return} $\bm{\pi}$\;}
	\Else{
	\For{$j=0,\dots, N_j$}{
	$\bm{\pi}\gets Replan (\bm{\pi},Z)$\;
	$\bm{\pi}\gets LSC(\bm{\pi})$\;
	\If{$RA(\bm{\pi},\Delta)$}{
	\textbf{return} $\bm{\pi}$\;}}

	$\bm{\pi}_{new}\gets \varnothing$\;
	\For{$i=0,\dots,size(\bm{\pi})-1$}{
		\If{$Steer(\bm{\pi}_i,\bm{\pi}_{i+1})$}{
			$\bm{\pi}_{\rm{new}}\gets\bm{\pi}_{\rm{new}}\cup\{\bm{\pi}_i,\bm{\pi}_{i+1}\}$}
		\Else{
			$\bm{\pi}_{\rm{local}}\gets\textit{RRT-SOS}(\bm{\pi}_i,\bm{\pi}_{i+1},\mathcal{X}_{\rm{obs}_i})$\;
			\If{$\bm{\pi}_{\rm{local}}$ is not None}{
			$\bm{\pi}_{\rm{new}} \gets \bm{\pi}_{\rm{local}} \cup \bm{\pi_{\rm{new}}}$\;}
		\Else{$\bm{\pi}_{\rm{new}}=\varnothing$}
		}
	}
	$\bm{\pi}\gets LSC(\bm{\pi}_{\rm{new}})$\;
	\If{$RA(\bm{\pi},\Delta)$}{
		\textbf{return} $\bm{\pi}$\;}
		\textbf{return} $\bm{\pi}=\varnothing$\;}
\end{algorithm}

\section{Simulation Experiments}
\label{sec:experiments}
In this section, we conduct several simulation experiments to demonstrate NR-RRT's distinguishing performance and generality.
We describe the collection of data and the implementation details of our neural network sampler in Section \ref{subsec:implements detail}.
Then, we compare our algorithm with RRT-SOS in uncertain environments with different acceptable risk levels in Section \ref{subsec:singal uncertain}.
In Section \ref{subsec:cluttered}, we further demonstrate that NR-RRT outperforms RRT-SOS in seen and unseen cluttered uncertain environments.

\subsection{Implementation Details}
\label{subsec:implements detail}
All experiments are conducted on the same system with $3.60\,\rm{GHz}\times8$ Intel Core i9-9900KF processor, 64GB RAM. 
Neural network models were trained with the PyTorch Python API on NVIDIA RTX 2080Ti.
The Adam optimizer \cite{kingma2014adam} is utilized with default parameters for training.
To ensure that our experiment is fair, we load the trained neural network model across the same CPUs as used in RRT-SOS algorithm.
The neural network only takes about $40$ms to predict the next state, which is a benefit to expanding the searching tree.

RRT-SOS algorithm is utilized to obtain expert demonstration paths.
Every expert path comprises several 2D points associated with the cost-to-go values with regard to their next points.
The $90\%$ of data is used for training, and the others become the validation dataset.
Besides, experiment maps are obtained by processing the risk contours map into images with the size of $256\times256$ pixels.
The batch size is 64, and the learning rate is fixed at $\eta=0.0004$.

\subsection{Path Planning in Uncertain Environments with Different Risk Levels}
\label{subsec:singal uncertain}
Although NR-RRT can handle diverse uncertainties, i.e., location, size, and geometry with various probability distributions, by constructing a risk contours map, we need to further prove its generality to different risk levels.
Thus, the objective of the following experiments is to find a near-optimal risk bounded path between a start and goal states in an uncertain environment with different risk levels $\Delta$.

We create two uncertain environments; one is convex and the other is nonconvex.
\begin{figure*}[htbp]
	\centering
		\subfigure[Planning Results in Uncertain Convex Environments with Different Risk Levels]{
		\includegraphics[width=18cm]{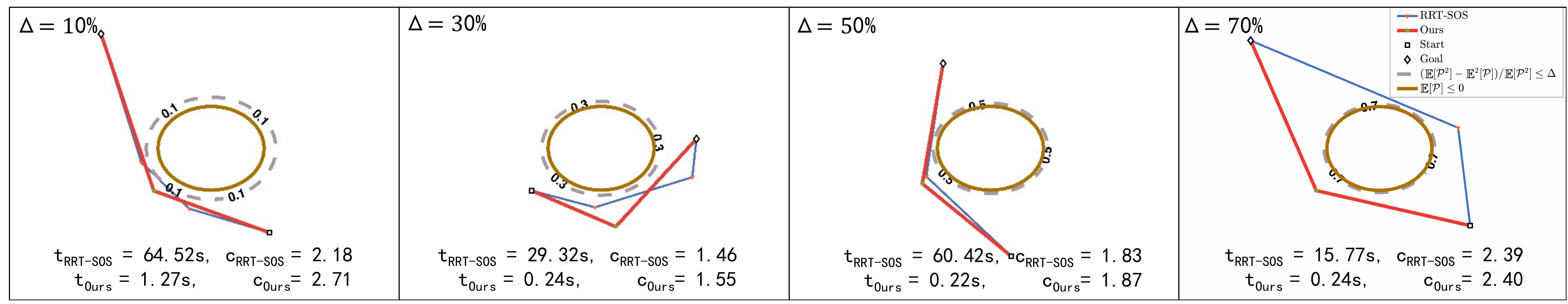}
		\label{fig:circle}
	}
	\subfigure[Planning Results in Uncertain Nonconvex Environments with Different Risk Levels]{
		\includegraphics[width=18cm]{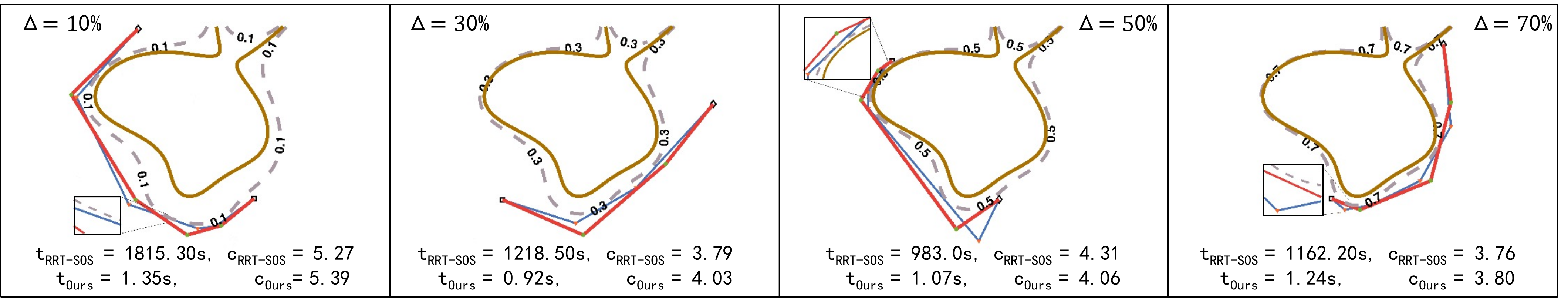}
		\label{fig:heart}
	}
	\caption{\small Figs. (a) and (b) are the comparisons of the planning results in uncertain convex and nonconvex environments with different risk levels, respectively. Given a risk-aware convex or nonconvex planning problem with some risk levels $\Delta$, NR-RRT (red) and RRT-SOS (blue) can find risk bounded low-cost path solutions that entirely do not lie in the dangerous and risk zone (brown and gray areas). NR-RRT uses extremely less time (\textbf{t}) than RRT-SOS takes to plan paths with comparable cost (\textbf{c}).}
	\label{trajectories}
\end{figure*}
The circle-shaped convex obstacle shown in Fig. \ref{fig:circle} is denoted as $\mathcal{X}_{\rm{obs}_1}(\omega)=\{(x,y):\omega^2-x^2-y^2\geq0\}$, where $(x,y)\in[-1,1]\times[-1,1]$, and the radius $\omega$ has a uniform probability distribution $\omega\sim\mathcal{U}(0.3,0.4)$. 
Note that this setup is the same as the illustrative example 3 in \cite{jasour2021convex} for comparison.
$\mathcal{X}_{\rm{obs}_2}(\omega)=\{(x,y):-0.35x^5-x^{4} y-0.5 x^{4}+0.2 x^{3} y^{2}-0.5 x^{3} y+0.31x^{3}-0.5 x^{2} y^{3}+0.2 x^2y^2+ 1.7 x^2y+0.26 x^2+0.7xy^4-0.1xy^{3}-1.5 x y^2-0.1 x y+ 0.1 x+0.02 y^{5}-0.1 y^{4}-0.04 y^{3}-0.1 y^{2}+0.28 y+0.89-0.7 \omega\}$ represents the heart-shaped nonconvex obstacle with uncertain parameter $\omega\sim {\mathrm{Beta}}(9,0.5)$, as shown in Fig. \ref{fig:heart}, where $(x,y)\in[-2,2]\times[-2,2]$.
We set ten risk levels $\Delta=\{10\%,20\%,\dots,100\%\}$ for each environment.
After choosing the risk level one by one for each environment, we randomly generate 1000 different valid start and goal pairs $(\bm{x}_{\rm{init}},\bm{x}_{\rm{goal}})$ for training, and randomly sample 200 pairs for testing.
Hence, we have 10000 training data and 2000 test data for every environment.

%\begin{table*}
%	\centering
%	\caption{Comparison of the Total Mean Planning Times and Path Lengths with Standard Deviations in Uncertain Convex and Nonconvex Environments With Ten Risk Levels}
%	\label{tab0}
%	\begin{tabular}{ccccc}
%		\toprule
%		&\multicolumn{2}{c}{Convex Env.} & \multicolumn{2}{c}{Nonconvex Env.} \\
%		\cmidrule(lr){2-3}   \cmidrule(lr){4-5}
%		&Time(s)&Length & Time(s)&Length \\
%		\hline
%		Ours  &$\bm{0.28\pm0.30 (0.26\pm0.11)}$  &$1.56\pm1.34 (2.17\pm0.88)$ & $\bm{1.08\pm0.28 (2.15\pm2.58)}$ & $4.06\pm0.24 (5.44\pm2.52)$ \\
%		\cite{jasour2021convex} &$21.30\pm36.67 (26.08\pm42.40)$  &$\bm{1.49\pm0.41 (1.60\pm0.78)}$&$1077.85\pm551.04 (1614.60\pm714.90)$ & $\bm{3.95\pm1.34 (4.22\pm1.23)}$\\
%		\bottomrule
%	\end{tabular}
%\end{table*}
\begin{table*}
	\centering
	\caption{Comparisons of the Total Mean Planning Times and Path Lengths with Standard Deviations in Uncertain Convex and Nonconvex Environments With Ten Risk Levels}
	\label{tab0}
	\begin{tabular}{ccccc}
		\toprule
		&\multicolumn{2}{c}{Convex Env.} & \multicolumn{2}{c}{Nonconvex Env.} \\
		\cmidrule(lr){2-3}   \cmidrule(lr){4-5}
		&Time(s)&Length & Time(s)&Length \\
		\hline
		Ours  &$\bm{0.26\pm0.11}$  &$2.17\pm0.88$ & $\bm{2.15\pm2.58}$ & $4.65\pm1.44$ \\
		\cite{jasour2021convex} &$26.08\pm42.40$  &$\bm{1.60\pm0.78}$&$1614.60\pm714.90$ & $\bm{4.22\pm1.23}$\\
		\bottomrule
	\end{tabular}
\end{table*}
The run-time to find the near-optimal solutions and their lengths are compared between two methods: RRT-SOS \cite{jasour2021convex} and NR-RRT. 
Table \ref{tab0} reports the total mean planning times and path lengths with standard deviations when the near-optimal trajectories are founded in uncertain environments with ten risk levels.
%\begin{table}[H]
%	\renewcommand{\arraystretch}{1.3}
%	\caption{Comparison of Algorithm Performance}
%	\label{tab1}
%	\centering
%	\begin{tabular}{cccc}
%		\toprule
%		& NPU RRT* & RRT-SOS & Time-varying SOS \\
%		\midrule
%		Found time   &                 &$6.542\pm2.496$-6.053 & \\
%		Planning time& $0.855\pm0.381$-0.723 &$59.439\pm21.434$-55.323 &\\
%		Path length  & $1.495\pm0.315$-1.407 &$1.489\pm0.314$-1.394&\\
%		\bottomrule
%	\end{tabular}
%\end{table}
We can see that our algorithm can generate near-optimal solutions with significantly less computation time than using RRT-SOS whether in the uncertain convex or nonconvex environment.
The medians of planning times are $27.12$s and $0.25$s by RRT-SOS and NR-RRT in the convex environment, respectively, while $1667.85$s by RRT-SOS and $0.93$s by NR-RRT in the nonconvex environment.
NR-RRT's smaller standard deviations and medians of the computation time imply it has a better stability.
Note that \textit{RA} takes about 0.09 seconds to verify the safety of every edge, which demonstrates that NR-RRT can be used in real-world applications.
Our method sometimes generates slightly shorter paths than the demonstrations generated by RRT-SOS because the expert paths are not always optimal, which will be discussed in Section \ref{sec:discuss}.
Therefore, no matter the risk level specified for an encountered uncertain obstacle, NR-RRT achieves a better performance than RRT-SOS by a large margin, thanks to the informative state sampling and bidirectional search strategy.

\subsection{Path Planning in Uncertain Cluttered Nonconvex Environments}
\label{subsec:cluttered}
\begin{figure}[htbp]
	\centering
	\includegraphics[width=0.48\textwidth]{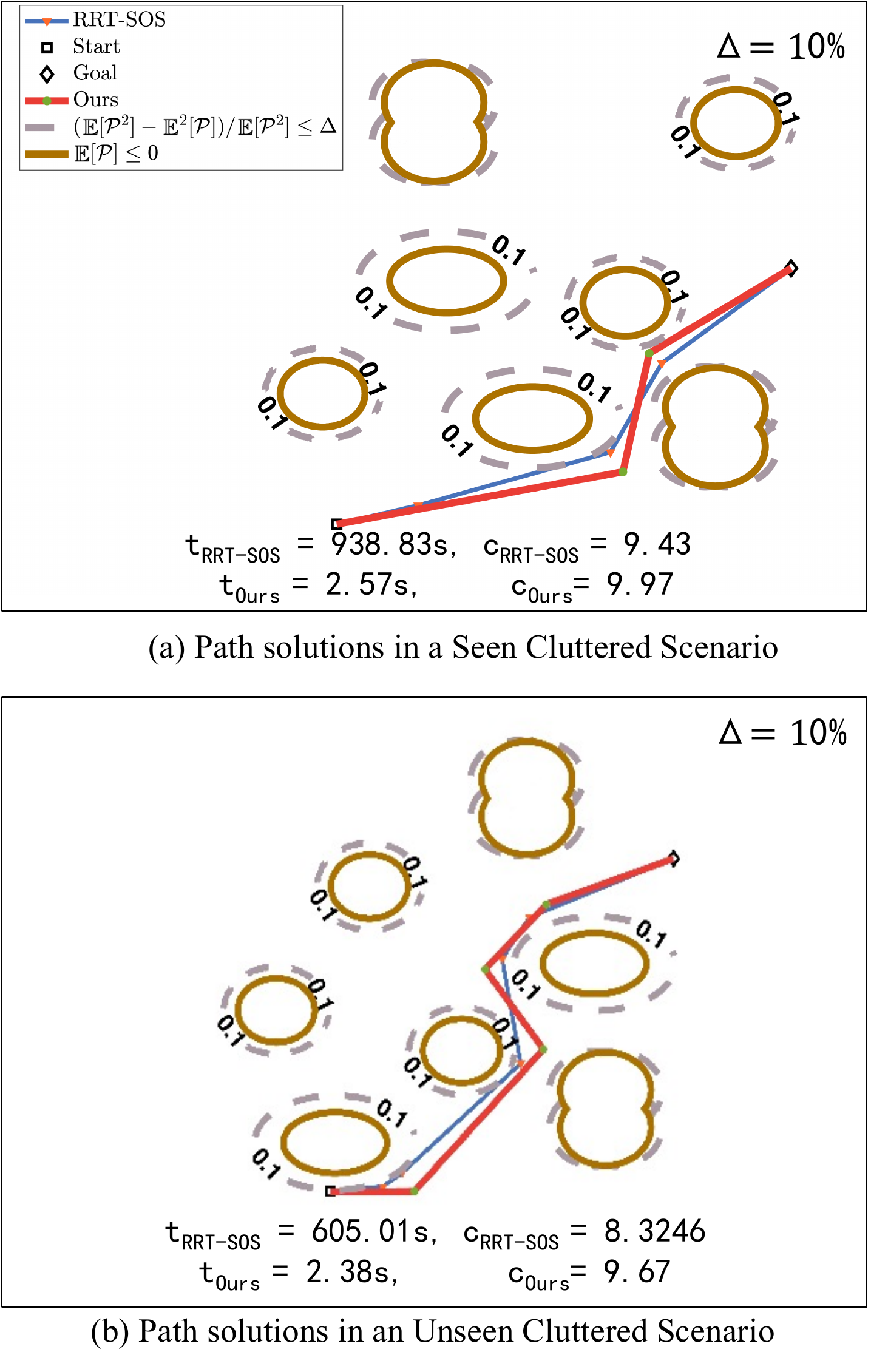}
	\caption{Comparisons of planning times (\textbf{t}) and path qualities (\textbf{c}) in seen (a) and unseen (b) uncertain cluttered nonconvex scenarios between NR-RRT (red) and RRT-SOS (blue) \cite{jasour2021convex}.}
	\label{fig:clutter2}
\end{figure}

In this section, we evaluate the performance of NR-RRT in seen and unseen uncertain cluttered nonconvex environments.
Twelve environments are created by random placement of seven uncertain obstacles.
Five are convex, i.e., three are circle-shaped and two are ellipse-shaped, and two are nonconvex and calabash-shaped.
The acceptable chance of collisions is set as $\Delta=10\%$.
We select ten of them as the seen environments for training and the remaining two as unseen scenarios only for the testing.
The range of the states is $[x,y]\in[-5,5]\times[-5,5]$.
We randomly sample 1000 different valid start and goal pairs in each seen environment.
As a result, we have 10000 training data belonging to ten seen environments.
On the other hand, we randomly create 200 start and goal pairs in every seen environment and 1000 pairs in every unseen environment.
Consequently, there are a total of 4000 data for the testing. 
Half of them are different pairs in the seen environments, while the others are new pairs in the unseen environments.
Figs. \ref{fig:clutter2} (a) and (b) show the planned trajectories in one seen and one unseen environment, respectively.
\begin{table*}
	\centering
	\caption{Comparisons of the Total Mean Planning Times and Path Lengths with Standard Deviations in Seen and Unseen Uncertain Cluttered Nonconvex Environments}
	\label{tab1}
	\begin{tabular}{ccccc}
		\toprule
		&\multicolumn{2}{c}{Seen Cluttered Env.} & \multicolumn{2}{c}{Unseen Cluttered Env.} \\
		\cmidrule(lr){2-3}   \cmidrule(lr){4-5}
		&Time(s)&Length & Time(s)&Length \\
		\hline
		Ours  &$\bm{2.69\pm0.28}$  & $7.40\pm2.40$ & $\bm{2.01\pm0.16}$ & $6.82\pm2.51$ \\
		\cite{jasour2021convex} &$313.48\pm669.53$  &$\bm{6.49\pm2.26}$&$237.90\pm518.50$ & $\bm{6.31\pm2.18}$\\
		\bottomrule
	\end{tabular}
\end{table*}

\begin{figure}[htbp]
	\centering
	\includegraphics[width=0.5\textwidth]{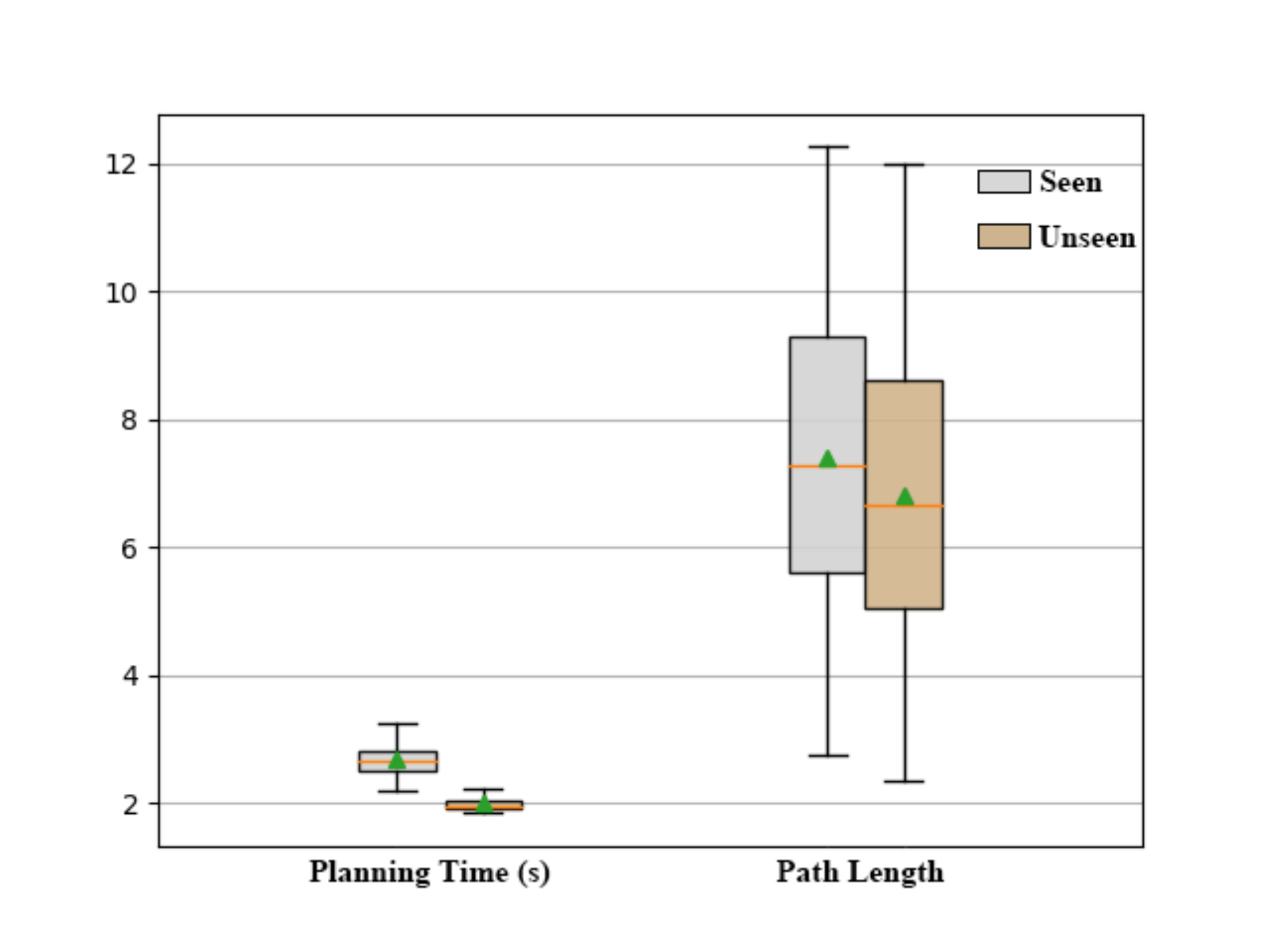}
	\caption{The interquartile ranges of planning times and path solution lengths for NR-RRT in seen and unseen uncertain cluttered nonconvex environments.}
	\label{fig:clutter2result}
\end{figure}

Table \ref{tab1} presents the comparisons of the total mean planning times and path lengths with standard deviations between NR-RRT and RRT-SOS \cite{jasour2021convex} when the near-optimal trajectories are founded in the ten seen and two unseen cluttered environments.
It can be seen that compared to RRT-SOS, NR-RRT takes a much shorter amount of time to look for similar quality trajectories in the seen and unseen environments.
Furthermore, Fig. \ref{fig:clutter2result} displays the box-plot of the planning times and path lengths that results from NR-RRT (for the sake of the large order of magnitudes, the box-plot for RRT-SOS is omitted).
The green triangle of the boxes refers to the average value, the orange line in the middle of the boxes is the median, and the height of the boxes refers to variance. 
We can tell that NR-RRT achieves high performance on the two indexes from this figure. 
The effectiveness of NR-RRT is also obvious on unseen tasks. 
Besides, in the given finite time interval, RRT-SOS and NR-RRT success rates are $100\%$.
The small variance of computation time indicates the robustness of NR-RRT in finding a near-optimal path in a short time.

\section{Discussion}
\label{sec:discuss}
In this section, NR-RRT's probabilistic completeness and optimality are analyzed.
\subsubsection{Probabilistic Completeness}
We first discuss the probabilistic completeness of NR-RRT, proposed as below:

\textbf{Proposition 1.} \textit{Given a risk-aware path planning problem $\left\{\bm{x}_{\rm{init}},\bm{x}_{\rm{goal}},\mathcal{X}_{\rm{obs}},\Delta\right\}$, and a risk assessor, NR-RRT can find a solution $\bm{\pi}$}: [0,T], \textit{if one exits, such that} $\bm{\pi}_0=\bm{x}_{\rm{init}},\,\bm{\pi}_T\in\mathcal{X}_{\rm{target}}$, \textit{and} $\text{Prob}(\bm{x}(t)\in\mathcal{X}_{\rm{obs}_i}(\omega_i))\leq\Delta|_{i=1}^{n_{o}}$.

Proposition 1 indicates that NR-RRT will find a risk bounded path, if one exists, where the probability of collision is no greater than $\Delta$.
Note that RRT-SOS is used as our standby SBMP that has \textit{probabilistic completeness}.
Based on the assumptions and Lemma as follows, we propose Theorem 1, which proves that the worst-case completeness guarantees of NR-RRT.
Therefore, the algorithm has probabilistic completeness as its standby planner RRT-SOS.

\textbf{Assumption 1.} \textit{The given start and goal states $(\bm{x}_{\rm{init}},\bm{x}_{{\rm{goal}}})$ are in safe space, i.e., $\bm{x}_{\rm{init}}\in\mathcal{X}_{\rm{safe}}$ and $\bm{x}_{{\rm{goal}}}\in\mathcal{X}_{\rm{target}}\subset\mathcal{X}_{\rm{safe}}$}. There exits at least one risk bounded path solution $\bm{\pi}$ that contains $\bm{x}_{\rm{init}}$ and $\bm{x}_{{\rm{goal}}}$, and $\bm{\pi}\not\subset\mathcal{X}_{\rm{obs}}$ by using a planner to find a risk bounded solution.

\textbf{Lemma 1.} \cite{qureshi2020motion} \textit{In a certain environment, if the start and goal pairs $(\bm{x}_{\rm{init}},\bm{x}_{{\rm{goal}}})$ is feasible, and there exists a collision-free path that connects them, for a planning problem $\left\{\bm{x}_{\rm{init}},\bm{x}_{{\rm{goal}}},\mathcal{X}_{obs}\right\}$, the probability of finding a solution approaches one as the underlying RRT* will be allowed to perform until infinity if needed for a iterative and recursive learning-based path planner.
}

\textbf{Theorem 1.} \textit{If Assumption 1 holds, for a risk-aware path planning problem $\left\{\bm{x}_{\rm{init}},\bm{x}_{{\rm{goal}}},\mathcal{X}_{obs},\Delta\right\}$, and a standby path planner whose sampler is with uniform distribution, NR-RRT will always find a risk bounded solution, if one exists. Because the planner that guarantees completeness is employed as a standby planner to run until infinity if NR-RRT fails to conclude a solution by using the neural network sampler for a ﬁxed number of trials.
}

\textit{Sketch of Proof:} 
\textit{Lemma 1} indicates that a neural planning method exhibits probabilistic completeness, which is with an underlying oracle planner, e.g., RRT* that guarantees completeness if one solution and a competent collision checker exist in an environment of certainty.
Similarly, assumption 1 states a condition that the risk-aware path planning problem is resolvable in our case.
Namely, at least one solution can be found by a planner that guarantees probabilistic completeness.
NR-RRT first obtains a coarse solution $\bm{\pi}$ that might contain non-connectable consecutive vertices.
Then, the algorithm tries to connect them via replanning.
These vertices lie in a safe space since each of them is evaluated by a risk assessor \textit{RA} before adding it to the vertex set of the solution $\bm{\pi}$.
During hybrid replanning, we aim to refine the coarse solution $\bm{\pi}$ by performing neural replanning.
Suppose there still exists any vertex that cannot be connected to conclude a final solution.
In that case, we use RRT-SOS to connect the remaining non-connectable vertices afterward.
The non-connectable vertices and the uncertain obstacles form a new risk-aware path planning problem and can be solved by RRT-SOS because of its probabilistic completeness.
Therefore, if Assumption 1 holds for the non-connectable vertices, NR-RRT guarantees the convergence of a solution inherited from the standby planner RRT-SOS.
NR-RRT with a standby planner RRT-SOS has probabilistic completeness in an uncertain environment.

\subsubsection{Optimality}
\label{sec:optimality}
RRT-SOS is employed to generate expert demonstrations. 
It also performs in the hybrid replanning and as a baseline against our NR-RRT.
However, the resulting paths using RRT-SOS are often non-optimal because the graph constructed by PRM only has a limited number of nodes in a map, and querying such a graph results in a non-global shortest solution.
In other words, RRT-SOS does not guarantee asymptotic optimality.
Since our resulting paths' quality cannot be the same as the training data, there is a situation where NR-RRT can find solutions with lower costs than the expert but non-optimal solutions.
Note that the ability to learn informed sampling distribution is more important than learning the exact positions of the labeled data for us.

In contrast, the number of samples in RRT* approaches infinity to obtain a graph covering the whole map area.
For RRT*, the probability of finding the optimal solution, if one exists, approaches one as the number of iterations increases to infinity. 
It gets asymptotic optimality from \textit{ChooseParent} and \textit{Rewire} processes that incrementally update the tree connections such that the shortest path is ensured.
Whenever RRT* attempts to connect $\bm{x}_{\rm{nearest}}$ from the $\bm{x}_{\rm{new}}$, it selects the best parent of $\bm{x}_{\rm{new}}$ in terms of the cost-to-come by searching the nodes in a certain radius in the \textit{ChooseParent}.
After adding a new edge about $\bm{x}_{\rm{new}}$ to the edge set, RRT* removes tree edges through $\bm{x}_{\rm{new}}$ with relatively higher costs, which is called \textit{Rewire}.
For more information, please refer to \cite{karaman2011sampling}.
Because our algorithm can naturally inherit asymptotic optimality from the standby planner, RRT-SOS, if it has, as proved in \cite{qureshi2020motion}, it is doable to modify the original RRT-SOS with these two processes to generate the optimal path.
However, there is no need to do this since the ability to find the optimal solution is not our contribution.
The resulting solutions found by NR-RRT have been near-optimal with little computational costs, which is more important in practice.
On the contrary, those modifications will significantly increase the computational time but offer inconspicuous improvement for the path quality, resulting in a low input-output ratio.

\section{Conclusions and Future Works}
\label{sec:conclu}
This article provides a learning-based bidirectional risk-aware path planning algorithm, NR-RRT, for quickly finding a risk bounded near-optimal solution in seen and unseen uncertain nonconvex environments where the obstacles may have probabilistic locations, sizes, and geometries.
The information of probabilistic nonconvex obstacles can be captured by constructing the risk contours map and further embedded in a deterministic latent feature map.
A proposed neural network sampler predicts the most-promising safe vertices after learning the past experiences. 
An informed bidirectional-search sampling strategy accelerates the convergence to a solution.
%A coarse path with guaranteed bounded risk is generated by conducting a bidirectional informed sampling strategy and refinement by hybrid replanning, which not only accelerates the convergence to a solution but guarantees the probabilistic completeness.
Our experiments show that our method can find comparable near-optimal solutions to the baseline but takes significantly less planning time in uncertain nonconvex environments.

In our future research directions, we will consider kinodynamic constraints (kinematics and differential constraints) \cite{li2016asymptotically} to make the planned trajectories easier to be tracked.
Besides, it is interesting to investigate path planning for robots under motion uncertainties \cite{agha2014firm}.

%\section*{Acknowledgment}

\ifCLASSOPTIONcaptionsoff
  \newpage
\fi



%\bibliographystyle{IEEEtran}
%\begin{thebibliography}{1}
%
%
%\end{thebibliography}
\bibliographystyle{IEEEtran}
\bibliography{./references.bib}

\begin{thebibliography}{10}
\providecommand{\url}[1]{#1}
\csname url@samestyle\endcsname
\providecommand{\newblock}{\relax}
\providecommand{\bibinfo}[2]{#2}
\providecommand{\BIBentrySTDinterwordspacing}{\spaceskip=0pt\relax}
\providecommand{\BIBentryALTinterwordstretchfactor}{4}
\providecommand{\BIBentryALTinterwordspacing}{\spaceskip=\fontdimen2\font plus
\BIBentryALTinterwordstretchfactor\fontdimen3\font minus
  \fontdimen4\font\relax}
\providecommand{\BIBforeignlanguage}[2]{{%
\expandafter\ifx\csname l@#1\endcsname\relax
\typeout{** WARNING: IEEEtran.bst: No hyphenation pattern has been}%
\typeout{** loaded for the language `#1'. Using the pattern for}%
\typeout{** the default language instead.}%
\else
\language=\csname l@#1\endcsname
\fi
#2}}
\providecommand{\BIBdecl}{\relax}
\BIBdecl

\bibitem{barbosa2021risk}
F.~S. Barbosa, B.~Lacerda, P.~Duckworth, J.~Tumova, and N.~Hawes, ``Risk-aware
  motion planning in partially known environments,'' \emph{arXiv preprint
  arXiv:2109.11287}, 2021.

\bibitem{da2019collision}
M.~da~Silva~Arantes, C.~F.~M. Toledo, B.~C. Williams, and M.~Ono,
  ``Collision-free encoding for chance-constrained nonconvex path planning,''
  \emph{IEEE Transactions on Robotics}, vol.~35, no.~2, pp. 433--448, 2019.

\bibitem{kavraki1996probabilistic}
L.~E. Kavraki, P.~Svestka, J.-C. Latombe, and M.~H. Overmars, ``Probabilistic
  roadmaps for path planning in high-dimensional configuration spaces,''
  \emph{IEEE transactions on Robotics and Automation}, vol.~12, no.~4, pp.
  566--580, 1996.

\bibitem{lavalle1998rapidly}
S.~M. LaValle \emph{et~al.}, ``Rapidly-exploring random trees: A new tool for
  path planning,'' 1998.

\bibitem{lavalle2006planning}
S.~M. LaValle, \emph{Planning algorithms}.\hskip 1em plus 0.5em minus
  0.4em\relax Cambridge university press, 2006.

\bibitem{jasour2021convex}
A.~Jasour, W.~Han, and B.~Williams, ``Convex risk bounded continuous-time
  trajectory planning in uncertain nonconvex environments,'' in \emph{Robotics:
  Science and Systems}, 2021.

\bibitem{karaman2011sampling}
S.~Karaman and E.~Frazzoli, ``Sampling-based algorithms for optimal motion
  planning,'' \emph{The international journal of robotics research}, vol.~30,
  no.~7, pp. 846--894, 2011.

\bibitem{axelrod2018provably}
B.~Axelrod, L.~P. Kaelbling, and T.~Lozano-P{\'e}rez, ``Provably safe robot
  navigation with obstacle uncertainty,'' \emph{The International Journal of
  Robotics Research}, vol.~37, no. 13-14, pp. 1760--1774, 2018.

\bibitem{luders2010chance}
B.~Luders, M.~Kothari, and J.~How, ``Chance constrained rrt for probabilistic
  robustness to environmental uncertainty,'' in \emph{AIAA guidance,
  navigation, and control conference}, 2010, p. 8160.

\bibitem{agha2014firm}
A.-A. Agha-Mohammadi, S.~Chakravorty, and N.~M. Amato, ``Firm: Sampling-based
  feedback motion-planning under motion uncertainty and imperfect
  measurements,'' \emph{The International Journal of Robotics Research},
  vol.~33, no.~2, pp. 268--304, 2014.

\bibitem{ho2022gaussian}
Q.~H. Ho, Z.~N. Sunberg, and M.~Lahijanian, ``Gaussian belief trees for chance
  constrained asymptotically optimal motion planning,'' \emph{arXiv preprint
  arXiv:2202.12407}, 2022.

\bibitem{jasour2019risk}
A.~M. Jasour and B.~C. Williams, ``Risk contours map for risk bounded motion
  planning under perception uncertainties.'' in \emph{Robotics: Science and
  Systems}, 2019.

\bibitem{wang2020neural}
J.~Wang, W.~Chi, C.~Li, C.~Wang, and M.~Q.-H. Meng, ``Neural rrt*:
  Learning-based optimal path planning,'' \emph{IEEE Transactions on Automation
  Science and Engineering}, vol.~17, no.~4, pp. 1748--1758, 2020.

\bibitem{hart1968formal}
P.~E. Hart, N.~J. Nilsson, and B.~Raphael, ``A formal basis for the heuristic
  determination of minimum cost paths,'' \emph{IEEE transactions on Systems
  Science and Cybernetics}, vol.~4, no.~2, pp. 100--107, 1968.

\bibitem{qureshi2020motion}
A.~H. Qureshi, Y.~Miao, A.~Simeonov, and M.~C. Yip, ``Motion planning networks:
  Bridging the gap between learning-based and classical motion planners,''
  \emph{IEEE Transactions on Robotics}, vol.~37, no.~1, pp. 48--66, 2020.

\bibitem{janson2018monte}
L.~Janson, E.~Schmerling, and M.~Pavone, ``Monte carlo motion planning for
  robot trajectory optimization under uncertainty,'' in \emph{Robotics
  Research}.\hskip 1em plus 0.5em minus 0.4em\relax Springer, 2018, pp.
  343--361.

\bibitem{cannon2017chance}
M.~Cannon, ``Chance-constrained optimization with tight confidence bounds,''
  \emph{arXiv preprint arXiv:1711.03747}, 2017.

\bibitem{blackmore2011chance}
L.~Blackmore, M.~Ono, and B.~C. Williams, ``Chance-constrained optimal path
  planning with obstacles,'' \emph{IEEE Transactions on Robotics}, vol.~27,
  no.~6, pp. 1080--1094, 2011.

\bibitem{johnson2021chance}
J.~J. Johnson and M.~C. Yip, ``Chance-constrained motion planning using modeled
  distance-to-collision functions,'' in \emph{2021 IEEE 17th International
  Conference on Automation Science and Engineering (CASE)}.\hskip 1em plus
  0.5em minus 0.4em\relax IEEE, 2021, pp. 1582--1589.

\bibitem{sun2016stochastic}
W.~Sun, J.~van~den Berg, and R.~Alterovitz, ``Stochastic extended lqr for
  optimization-based motion planning under uncertainty,'' \emph{IEEE
  Transactions on Automation Science and Engineering}, vol.~13, no.~2, pp.
  437--447, 2016.

\bibitem{sun2015high}
W.~Sun, S.~Patil, and R.~Alterovitz, ``High-frequency replanning under
  uncertainty using parallel sampling-based motion planning,'' \emph{IEEE
  Transactions on Robotics}, vol.~31, no.~1, pp. 104--116, 2015.

\bibitem{van2011lqg}
J.~Van Den~Berg, P.~Abbeel, and K.~Goldberg, ``Lqg-mp: Optimized path planning
  for robots with motion uncertainty and imperfect state information,''
  \emph{The International Journal of Robotics Research}, vol.~30, no.~7, pp.
  895--913, 2011.

\bibitem{jaillet2011eg}
L.~Jaillet, J.~Hoffman, J.~Van~den Berg, P.~Abbeel, J.~M. Porta, and
  K.~Goldberg, ``Eg-rrt: Environment-guided random trees for kinodynamic motion
  planning with uncertainty and obstacles,'' in \emph{2011 IEEE/RSJ
  International Conference on Intelligent Robots and Systems}.\hskip 1em plus
  0.5em minus 0.4em\relax IEEE, 2011, pp. 2646--2652.

\bibitem{alterovitz2007stochastic}
R.~Alterovitz, T.~Sim{\'e}on, and K.~Goldberg, ``The stochastic motion roadmap:
  A sampling framework for planning with markov motion uncertainty,'' in
  \emph{Robotics: Science and systems}, 2007.

\bibitem{van2012motion}
J.~Van Den~Berg, S.~Patil, and R.~Alterovitz, ``Motion planning under
  uncertainty using iterative local optimization in belief space,'' \emph{The
  International Journal of Robotics Research}, vol.~31, no.~11, pp. 1263--1278,
  2012.

\bibitem{cai2021hyp}
P.~Cai, Y.~Luo, D.~Hsu, and W.~S. Lee, ``Hyp-despot: A hybrid parallel
  algorithm for online planning under uncertainty,'' \emph{The International
  Journal of Robotics Research}, vol.~40, no. 2-3, pp. 558--573, 2021.

\bibitem{lee2020magic}
Y.~Lee, P.~Cai, and D.~Hsu, ``Magic: Learning macro-actions for online pomdp
  planning,'' \emph{arXiv preprint arXiv:2011.03813}, 2020.

\bibitem{richards2002aircraft}
A.~Richards and J.~P. How, ``Aircraft trajectory planning with collision
  avoidance using mixed integer linear programming,'' in \emph{Proceedings of
  the 2002 American Control Conference (IEEE Cat. No. CH37301)}, vol.~3.\hskip
  1em plus 0.5em minus 0.4em\relax IEEE, 2002, pp. 1936--1941.

\bibitem{summers2018distributionally}
T.~Summers, ``Distributionally robust sampling-based motion planning under
  uncertainty,'' in \emph{2018 IEEE/RSJ International Conference on Intelligent
  Robots and Systems (IROS)}.\hskip 1em plus 0.5em minus 0.4em\relax IEEE,
  2018, pp. 6518--6523.

\bibitem{https://doi.org/10.1049/csy2.12020}
\BIBentryALTinterwordspacing
J.~Wang, T.~Zhang, N.~Ma, Z.~Li, H.~Ma, F.~Meng, and M.~Q.-H. Meng, ``A survey
  of learning-based robot motion planning,'' \emph{IET Cyber-Systems and
  Robotics}, vol.~3, no.~4, pp. 302--314, 2021. [Online]. Available:
  \url{https://ietresearch.onlinelibrary.wiley.com/doi/abs/10.1049/csy2.12020}
\BIBentrySTDinterwordspacing

\bibitem{srinivas2018universal}
A.~Srinivas, A.~Jabri, P.~Abbeel, S.~Levine, and C.~Finn, ``Universal planning
  networks: Learning generalizable representations for visuomotor control,'' in
  \emph{International Conference on Machine Learning}.\hskip 1em plus 0.5em
  minus 0.4em\relax PMLR, 2018, pp. 4732--4741.

\bibitem{tamar2016value}
A.~Tamar, Y.~Wu, G.~Thomas, S.~Levine, and P.~Abbeel, ``Value iteration
  networks,'' \emph{Advances in neural information processing systems},
  vol.~29, 2016.

\bibitem{faust2018prm}
A.~Faust, K.~Oslund, O.~Ramirez, A.~Francis, L.~Tapia, M.~Fiser, and
  J.~Davidson, ``Prm-rl: Long-range robotic navigation tasks by combining
  reinforcement learning and sampling-based planning,'' in \emph{2018 IEEE
  International Conference on Robotics and Automation (ICRA)}.\hskip 1em plus
  0.5em minus 0.4em\relax IEEE, 2018, pp. 5113--5120.

\bibitem{fan2020learning}
T.~Fan, P.~Long, W.~Liu, J.~Pan, R.~Yang, and D.~Manocha, ``Learning resilient
  behaviors for navigation under uncertainty,'' in \emph{2020 IEEE
  International Conference on Robotics and Automation (ICRA)}.\hskip 1em plus
  0.5em minus 0.4em\relax IEEE, 2020, pp. 5299--5305.

\bibitem{ichter2018learning}
B.~Ichter, J.~Harrison, and M.~Pavone, ``Learning sampling distributions for
  robot motion planning,'' in \emph{2018 IEEE International Conference on
  Robotics and Automation (ICRA)}.\hskip 1em plus 0.5em minus 0.4em\relax IEEE,
  2018, pp. 7087--7094.

\bibitem{ichter2019robot}
B.~Ichter and M.~Pavone, ``Robot motion planning in learned latent spaces,''
  \emph{IEEE Robotics and Automation Letters}, vol.~4, no.~3, pp. 2407--2414,
  2019.

\bibitem{ma2022enhanced}
\BIBentryALTinterwordspacing
H.~Ma, C.~Li, J.~Liu, J.~Wang, and M.~Q. Meng, ``Enhance connectivity of
  promising regions for sampling-based path planning,'' \emph{CoRR}, vol.
  abs/2112.08106, 2021. [Online]. Available:
  \url{https://arxiv.org/abs/2112.08106}
\BIBentrySTDinterwordspacing

\bibitem{zhang2021generative}
T.~Zhang, J.~Wang, and M.~Q.-H. Meng, ``Generative adversarial network based
  heuristics for sampling-based path planning,'' \emph{IEEE/CAA Journal of
  Automatica Sinica}, vol.~9, no.~1, pp. 64--74, 2021.

\bibitem{girshick2015fast}
R.~Girshick, ``Fast r-cnn,'' in \emph{Proceedings of the IEEE international
  conference on computer vision}, 2015, pp. 1440--1448.

\bibitem{srivastava2014dropout}
N.~Srivastava, G.~Hinton, A.~Krizhevsky, I.~Sutskever, and R.~Salakhutdinov,
  ``Dropout: a simple way to prevent neural networks from overfitting,''
  \emph{The journal of machine learning research}, vol.~15, no.~1, pp.
  1929--1958, 2014.

\bibitem{tobenkin2013spotless}
M.~M. Tobenkin, F.~Permenter, and A.~Megretski, ``Spotless polynomial and conic
  optimization,'' \emph{View online}, 2013.

\bibitem{hauser2010fast}
K.~Hauser and V.~Ng-Thow-Hing, ``Fast smoothing of manipulator trajectories
  using optimal bounded-acceleration shortcuts,'' in \emph{2010 IEEE
  international conference on robotics and automation}.\hskip 1em plus 0.5em
  minus 0.4em\relax IEEE, 2010, pp. 2493--2498.

\bibitem{kingma2014adam}
D.~P. Kingma and J.~Ba, ``Adam: A method for stochastic optimization,''
  \emph{arXiv preprint arXiv:1412.6980}, 2014.

\bibitem{li2016asymptotically}
Y.~Li, Z.~Littlefield, and K.~E. Bekris, ``Asymptotically optimal
  sampling-based kinodynamic planning,'' \emph{The International Journal of
  Robotics Research}, vol.~35, no.~5, pp. 528--564, 2016.

\end{thebibliography}

\begin{IEEEbiography}[{\includegraphics[width=1in,height=1.25in,clip,keepaspectratio]{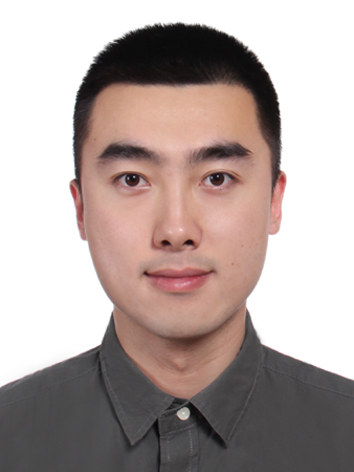}}]{Fei Meng}
received the B.Eng. in electrical engineering and automation, and the M.Eng. in control engineering from Harbin Institute of Technology, Weihai and Harbin, China, in 2016 and 2019, respectively. He is working toward the Ph.D. degree with the Department of Electronic Engineering, The Chinese University of Hong Kong, Hong Kong SAR, China. 

His research interests include learning-based motion planning and control.
\end{IEEEbiography}

\begin{IEEEbiography}[{\includegraphics[width=1in,height=1.25in,clip,keepaspectratio]{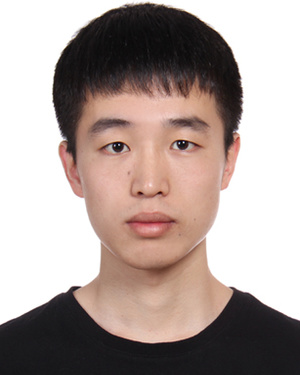}}]{Liangliang Chen}
	received the B.B.A. degree in business administration in 2017, the B.S. degree in automation in 2017, and M.Eng. degree in control engineering in 2019, all from Harbin Institute of Technology, Harbin, China. He is currently pursuing the Ph.D. degree in electrical and computer engineering at Georgia Institute of Technology, Atlanta, GA, USA.
	
	His current research interests include deep reinforcement learning.
\end{IEEEbiography}

%\newpage

\begin{IEEEbiography}[{\includegraphics[width=1in,height=1.25in,clip,keepaspectratio]{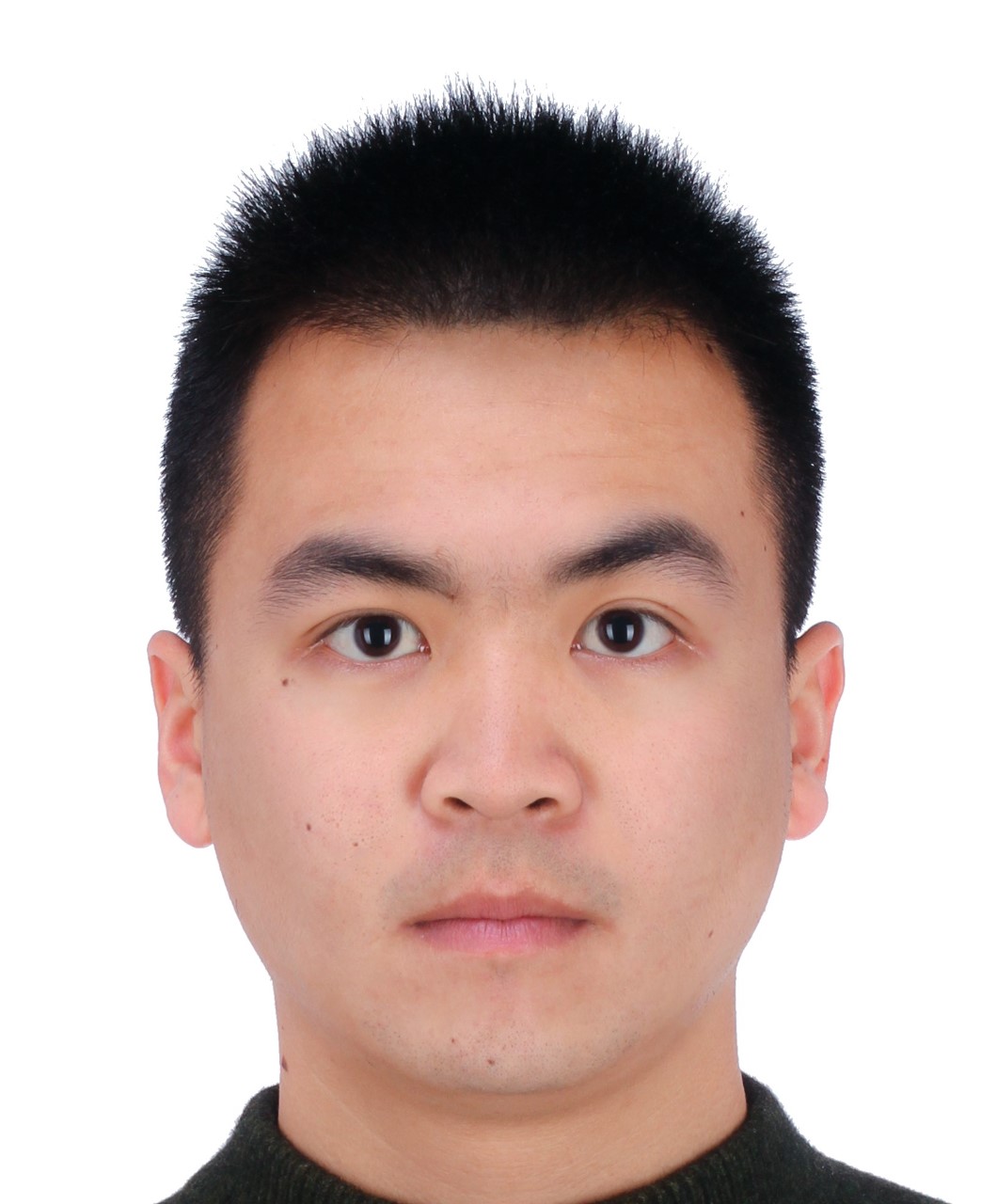}}]{Han Ma}
received the B.E. degree in measurement, control technology and instrument from the Department of Precision Instrument of Tsinghua University, Beijing, China, in 2019. He is now working towards the Ph.D. degree in the Department of Electronic Engineering of The Chinese University of Hong Kong, Hong Kong SAR, China. 

His research interests include path planning and machine learning in robotics.
\end{IEEEbiography}

\begin{IEEEbiography}[{\includegraphics[width=1in,height=1.25in,clip,keepaspectratio]{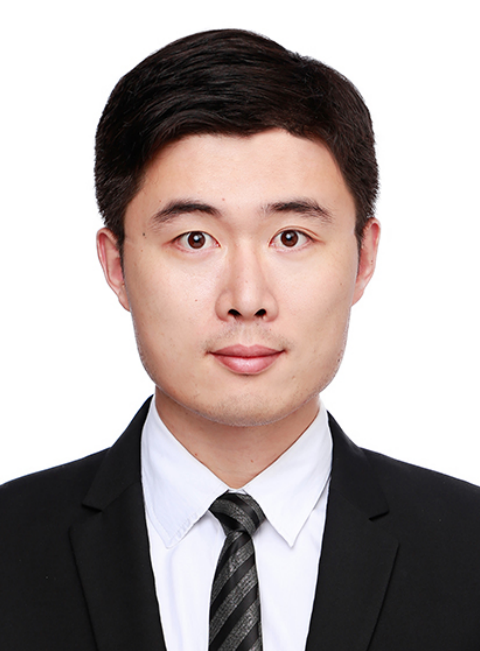}}]{Jiankun Wang}
received the B.E. degree in Automation from Shandong University, Jinan, China, in 2015, and the Ph.D. degree in Department of Electronic Engineering, The Chinese University of Hong Kong, Hong Kong, in 2019. He is currently a Research Assistant Professor with the Department of Electronic and Electrical Engineering of the Southern University of Science and Technology, Shenzhen, China.

During his Ph.D. degree, he spent six months at Stanford University, CA, USA, as a Visiting Student Scholar supervised by Prof. Oussama Khatib. His current research interests include motion planning and control, human robot interaction, and machine learning in robotics.
\end{IEEEbiography}

\begin{IEEEbiography}[{\includegraphics[width=1in,height=1.25in,clip,keepaspectratio]{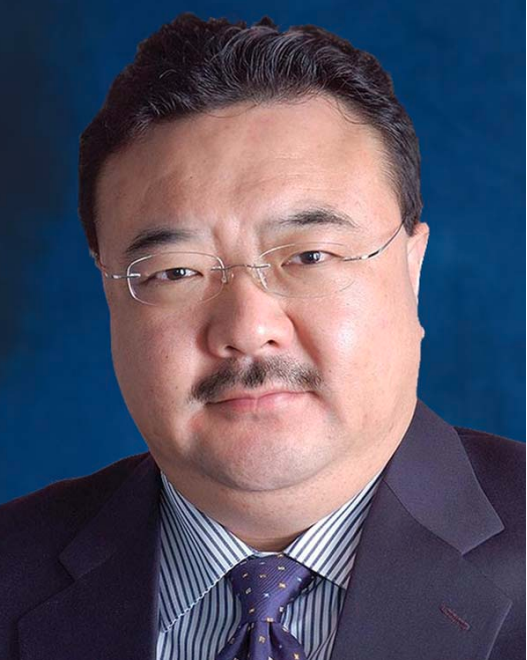}}]{Max Q.-H. Meng} received his Ph.D. degree in Electrical and Computer Engineering from the University of Victoria, Canada, in 1992. He is currently a Chair Professor and the Head of the Department of Electronic and Electrical Engineering at the Southern University of Science and Technology in Shenzhen, China, on leave from the Department of Electronic Engineering at the Chinese University of Hong Kong. He joined the Chinese University of Hong Kong in 2001 as a Professor and later the Chairman of Department of Electronic Engineering. He was with the Department of Electrical and Computer Engineering at the University of Alberta in Canada, where he served as the Director of the ART (Advanced Robotics and Teleoperation) Lab and held the positions of Assistant Professor (1994), Associate Professor (1998), and Professor (2000), respectively. He is an Honorary Chair Professor at Harbin Institute of Technology and Zhejiang University, and also the Honorary Dean of the School of Control Science and Engineering at Shandong University, in China. 
	His research interests include medical and service robotics, robotics perception and intelligence. He has published more than 750 journal and conference papers and book chapters and led more than 60 funded research projects to completion as Principal Investigator. 
	Prof. Meng has been serving as the Editor-in-Chief and editorial board of a number of international journals, including the Editor-in-Chief of the Elsevier Journal of Biomimetic Intelligence and Robotics, and as the General Chair or Program Chair of many international conferences, including the General Chair of IROS 2005 and ICRA 2021, respectively. He served as an Associate VP for Conferences of the IEEE Robotics and Automation Society (2004-2007), Co-Chair of the Fellow Evaluation Committee and an elected member of the AdCom of IEEE RAS for two terms. He is a recipient of the IEEE Millennium Medal, a Fellow of IEEE, a Fellow of Hong Kong Institution of Engineers, and an Academician of the Canadian Academy of Engineering.
\end{IEEEbiography}


%\vfill


\end{document}